\documentclass[sigconf]{acmart}
\DeclareUnicodeCharacter{FF0C}{,}
\usepackage{subfigure}
\usepackage{threeparttable}
\usepackage[ruled,lined,linesnumbered]{algorithm2e}

\def\model{\textsc{PHER}}

\AtBeginDocument{%
  }

\copyrightyear{2026}
\acmYear{2026}
\setcopyright{cc}
\setcctype{by}
\acmConference[CIKM '26]{Proceedings of the 35th ACM International Conference on Information and Knowledge Management}{November 07--11, 2026}{Rome, Italy}
\acmBooktitle{Proceedings of the 35th ACM International Conference on Information and Knowledge Management (CIKM '26), November 07--11, 2026, Rome, Italy}
\acmDOI{10.1145/3799682.3840882}
\acmISBN{979-8-4007-2539-5/2026/11}

\begin{document}

%remove ACM format information
% \settopmatter{printfolios=true}
% \settopmatter{printacmref=true} % Removes citation information below abstract
% \renewcommand\footnotetextcopyrightpermission[1]{} % removes footnote with conference information in first column
% \pagestyle{empty} % removes running headers
% \title{Continuous Optimization for Feature Selection with Policy-Guided Search and Autoregressive Reconstruction}
% \title{Permutation-Invariant Hierarchical Modeling for Reinforcement-Enhanced Generative Feature Transformation}

\title{Hierarchical and Permutation-Invariant Feature Transformation Learning via Policy-Guided Embedding Search}

%% author information
\author{Rui Liu}
\affiliation{
  \institution{University of Kansas}
  \city{Lawrence}
  \country{United States}}
\email{rayliu@ku.edu}

\author{Tao Zhe}
\affiliation{
  \institution{University of Kansas}
  \city{Lawrence}
  \country{United States}}
\email{taozhe@ku.edu}

\author{Yanyong Huang$^{\dag}$}
\affiliation{
  \institution{Southwest University of Finance and Economics}
  \city{Chengdu}
  \country{China}}
\email{huangyy@swufe.edu.cn}

\author{Sankha Narayan Guria}
\affiliation{
  \institution{University of Kansas}
  \city{Lawrence}
  \country{United States}}
\email{sankha@ku.edu}

\author{Xiao Luo}
\affiliation{
  \institution{University of Wisconsin--Madison}
  \city{Madison}
  \country{United States}}
\email{xiao.luo@wisc.edu}

\author{Wei Fan}
\affiliation{
  \institution{University of Auckland}
  \city{Auckland}
  \country{New Zealand}}
\email{wei.fan@auckland.ac.nz}

\author{Yanjie Fu}
\affiliation{
  \institution{Arizona State University}
  \city{Tempe}
  \country{United States}}
\email{yanjie.fu@asu.edu}

\author{Dongjie Wang$^{\dag}$}
\authornote{Corresponding authors.}
\affiliation{
  \institution{Northeast Normal University}
  \city{Jilin}
  \country{China}}
\email{wangdongjie100@nenu.edu.cn}

\renewcommand{\shortauthors}{Rui Liu et al.}

\begin{abstract}
Feature transformation improves predictive performance on tabular data by constructing informative abstractions from raw features.
Recent generative approaches encode transformation knowledge into continuous embedding spaces for efficient exploration of candidate strategies, but face three key limitations:
1) overlooking hierarchical relationships between low-level features, operations, and high-level abstractions;
2) enforcing order-sensitive embeddings on inherently permutation-invariant transformation sequences, introducing systematic bias;
and 3) relying on gradient-based search ill-suited to non-convex transformation spaces.
We propose a framework with two complementary components.
First, a permutation-invariant hierarchical module captures interactions across features, operations, and abstraction levels, with a self-attention pooling mechanism that maps semantically equivalent structures to consistent embeddings aligned with downstream performance.
Second, a policy-guided multi-objective reinforcement learning strategy initializes search from empirically strong seeds and jointly optimizes predictive accuracy and transformation efficiency.
Extensive experiments on diverse tabular benchmarks demonstrate the effectiveness and robustness of our framework against strong baselines.
Our code and data are publicly available
\footnote{\url{https://github.com/RayLiu1103/PHER}}.
\end{abstract}
% \ccsdesc[500]{Information systems~Data mining}




\maketitle

% \vspace{-0.3cm}
\section{Introduction}
Feature transformation aims to improve tabular feature spaces by mathematically deriving more predictive representations from original features.
Although deep learning has recently achieved remarkable success, it struggles to deliver strong performance on tabular data due to heterogeneous feature types, varying feature scales, and the presence of missing values or outliers. 
Furthermore, the black-box nature of deep learning models limits their interpretability, hindering their application in further data analysis and critical decision-making systems.
Thus, automated feature transformation, which aims to generate distinguishable and informative features to enhance predictive performance, has emerged as a significant research direction in tabular data analysis.

Existing methods for automated feature transformation can be categorized into three main groups:
1) Expansion-reduction approaches~\citep{kanter2015deep,afat,khurana2016cognito}, which first enlarge the original feature space using mathematical transformations and subsequently reduce dimensionality through feature selection;
2) Evolution-evaluation approaches~\citep{grfg,ttg,tran2016genetic}, which utilize Evolutionary Algorithms (EA) or Reinforcement Learning (RL) to iteratively generate and evaluate transformation candidates based on model performance feedback;
3) Auto ML-based approaches~\citep{nfs,difer}, which formulate feature transformation as a search problem, employing automated machine learning techniques to identify optimal transformations.
Despite achieving considerable success, these methods still face challenges in effectively modeling intricate patterns inherent in feature transformation knowledge.

Inspired by the success of generative AI, recent advances formulate automated feature transformation as a token generation task~\citep{moat,ying2024unsupervised,ying2024feature,caps,liu2026permutationinvariantrepresentationlearningrobust}.
They compress feature transformation knowledge into a continuous embedding space and subsequently explore this space via gradient-based methods to identify superior feature transformation sequences.
However, there are three primary limitations:
1) \textbf{Neglecting hierarchical relationships between original features, mathematical operations, and generated feature abstractions.}
\begin{figure}
% \vspace{-0.4cm}
  \begin{center}
  \includegraphics[width=0.5\textwidth]{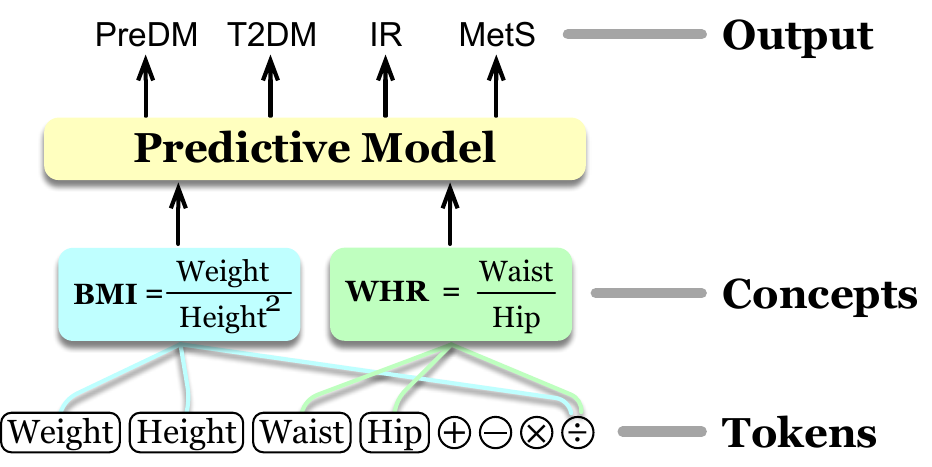}
  \end{center}
  % \vspace{-0.5cm}
\caption{Example of hierarchical relationships between generated feature abstractions (concepts) and original features. BMI and WHR are derived from original features.}

  \label{motivation}
  % \vspace{-0.45cm}
\end{figure}
For example, as illustrated in Figure~\ref{motivation}, Body Mass Index (BMI) and Waist-to-Hip Ratio (WHR) are high-level feature abstractions (i.e., concepts) derived from original features through mathematical transformations, which can significantly contribute to predicting clinical outcomes such as Pre-Diabetes (PreDM), Type~2 Diabetes (T2DM), Insulin Resistance (IR), and Metabolic Syndrome (MetS). 
Neglecting explicit modeling of these hierarchical relationships results in incomplete feature representations, limiting the capture of meaningful interactions and reducing the discriminative power of the embedding space.
2) \textbf{Ignoring the order invariance of the generated feature abstractions with respect to the predictive performance of the associated feature space.}
Prior studies treat the entire feature transformation process as a sequential model to capture transformation knowledge.
However, this formulation introduces unnecessary permutation bias among generated concepts into the embedding space, hindering effective exploration and limiting the identification of the globally optimal feature transformation embedding.
3) \textbf{Relying on gradient-based search methods that strongly assume convexity of the learned embedding space.}
Existing methods typically assume convexity in the learned embedding space, relying on gradient-based search methods to identify globally optimal solutions.
However, due to complex interactions among features and mathematical operations, it is challenging to guarantee the convexity of the embedding space.
This misalignment between the convexity assumption and the practical characteristics of the embedding space increases the risk of being trapped in local optima, resulting in suboptimal feature transformation sequences.

\textbf{Our Contribution: A Hierarchical Modeling and Policy-Guided Feature Transformation Perspective.}
To address these limitations, we propose {\model}, a novel feature transformation framework that integrates permutation-invariant hierarchical modeling and multi-objective policy-guided search.
Specifically, given a large volume of feature transformation records, where each record consists of a transformation sequence and the associated model performance, we first design a hierarchical modeling module. 
This module preserves transformation knowledge at both low-level feature interactions and high-level feature abstractions (i.e. \textit{concepts}) into a continuous embedding space.
To ensure permutation invariance, we develop a self-attention pooling mechanism that symmetrically computes attention scores across various generated concepts.
This structure guarantees that any permutation of generated concepts yields identical embeddings, resulting in a global continuous embedding space that unbiasedly preserves feature transformation knowledge.
Subsequently, we employ a policy-guided multi-objective search strategy to explore the global embedding space and identify the optimal feature transformation sequence.
In detail, we first select the top-K feature transformation sequences based on model performance as search seeds.
Then, we convert these seeds into embeddings, which serve as initial positions for exploration within the learned global embedding space.
Next, we treat seed embeddings as states and implement a reinforcement learning agent to explore the embedding space, guided by maximizing downstream task performance and minimizing the length of transformation sequences.
The exploratory nature of reinforcement learning enables effective navigation of the embedding space, reducing the risk of becoming trapped in local optima, even when the embedding space lacks convexity.
Finally, we conduct extensive experiments on 19 real-world datasets to evaluate the efficiency, resilience, and traceability of \model.
The experimental results show the superior performance of {\model} over the state-of-the-art feature transformation methods.

\begin{figure*}[!thbp]
% \vspace{-0.3cm}
    \centering
    \includegraphics[width=\linewidth]{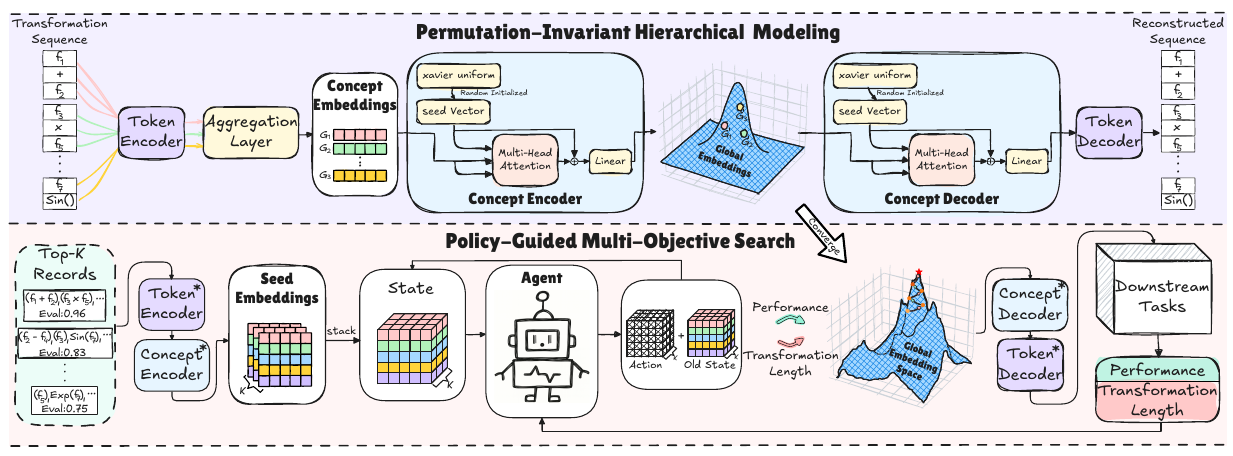}
    % \includegraphics[width=\linewidth]{fig/1.pdf}
    % \vspace{-0.7cm}
    \caption{An overview of our framework. {\model} comprises two main components:
    1) Permutation-Invariant Hierarchical Modeling, which unbiasedly preserves feature transformation knowledge at both the feature-operation token level and the generated-concept level within a global embedding space;
    2) Policy-Guided Multi-objective Search, which effectively explores the learned embedding space to identify optimal feature transformation sequences.}
    \label{model_overview}
    \vspace{0.4cm}
\end{figure*}

\vspace{-0.2cm}
\section{Problem Statement}
We aim to develop an automated feature transformation framework from a generative intelligence perspective, integrating permutation-invariant hierarchical modeling and a policy-guided multi-objective search strategy.
Formally, given a dataset $D=\{X,y\}$, where $X$ denotes features and $y$ represents the corresponding labels, along with a mathematical operation set $\mathcal{O}$ (e.g., addition, subtraction, multiplication). 
We first gather $n$ feature transformation records, denoted by ${\{(\mathbf{\Gamma_{i}},v_{i})\}}^{n}_{i=1}$ based on the dataset $D$,  where each record consists of a transformation sequence $\mathbf{\Gamma_{i}}$ and the associated downstream predictive performance $v_i$.
We then train a token-level encoder $\phi_{tok}$ and decoder $\psi_{tok}$ to encode interactions between original features and mathematical operations into token embeddings $\mathbf{E}$, optimized through a token-level reconstruction loss.
These token embeddings are grouped and averaged through an aggregation layer to form concept embeddings $\mathbf{G}$.
A concept-level permutation-invariant encoder
$\phi_{con}$ is then employed to eliminate order sensitivity among concepts, resulting in a global, unbiased embedding space $\mathbb{G}$.
A corresponding concept-level decoder $\psi_{con}$ is simultaneously trained to capture patterns from generated concepts.
After that, we employ a policy-guided strategy to explore $\mathbb{G}$, aiming to identify the optimal global embedding $\mathbf{G}^{'}_{opt}$. 
This embedding can be decoded by the concept-level decoder $\psi_{con}$ and token-level decoder $\psi_{tok}$ to reconstruct the optimal feature transformation sequence $\mathbf{\Gamma^{*}}$, thus generating a feature space that maximizes downstream task performance $\mathcal{M}$.
% Appendix~\ref{appendix:notations} summarizes all mathematical notations used in this paper.
Formally, the optimization goal is formulated as:
\begin{equation}
    \mathbf{\Gamma^{*}} = \psi_{tok}(\psi_{con}(\mathbf{G}^{'}_{opt}))=\mathrm{argmax}_{\mathbf{G}^{'}\in\mathbb{G}}\mathcal{M}(X[\psi_{tok}(\psi_{con}(\mathbf{G}^{'}))]).
\end{equation}

% \vspace{-0.5cm}
\section{Methodology}
\subsection{Framework Overview}
Figure~\ref{model_overview} illustrates the overall structure of \model, comprising two primary components: 1) Permutation-Invariant Hierarchical Modeling and 2) Policy-Guided Multi-Objective Search. Specifically, given a large collection of feature transformation records, each record consists of one feature transformation sequence and the corresponding model performance.
We first develop a token-level encoder to embed original features and mathematical operations, capturing token-level transformation patterns. 
Next, we aggregate token embeddings into concept embeddings according to the correspondence between original features, operations, and generated concepts. 
We then apply a concept-level encoder with a self-attention pooling mechanism to eliminate order sensitivity, obtaining permutation-invariant global embeddings. 
To effectively train the encoders, we introduce a concept-level decoder and a token-level decoder to reconstruct concept embeddings and original feature-operation tokens, respectively.
After obtaining the global embedding space, we employ a policy-guided multi-objective search strategy to explore this embedding space, guided by maximizing downstream task performance and minimizing the length of feature transformation sequences.
The identified optimal embedding is then decoded into its corresponding feature transformation sequence using the trained decoder. Finally, this reconstructed sequence is applied to the original feature space, resulting in an optimized feature space with improved predictive performance.

% \vspace{-0.2cm}
\subsection{Hierarchical Feature Transformation Knowledge Modeling}
\noindent\textbf{Why Hierarchical Knowledge Modeling Matters.}
Feature transformation inherently involves intricate hierarchical relationships among original features, mathematical operations, and higher-level generated concepts. 
However, existing generative intelligence-based approaches typically treat the entire feature transformation sequence as an indistinguishable whole, neglecting these hierarchical distinctions. 
This oversimplification leads to inadequate modeling of transformation knowledge, reducing the discriminative power of the continuous embedding space, and ultimately resulting in suboptimal feature spaces. 
To address these limitations, we develop a hierarchical modeling module in our framework, which can explicitly capture both token-level relationships (individual feature-operation interactions) and concept-level relationships (aggregated higher-level concepts) within the learned embedding space.

\noindent\textbf{Reinforcement Transformation-Accuracy Training Data Collection.}
To learn an effective embedding space of feature transformation, we collect $n$ transformation records using an RL-based method~\citep{grfg}, denoted by ${\{(\mathbf{\Gamma_{i}},v_{i})\}}^{n}_{i=1}$, where $\mathbf{\Gamma_{i}}$ is a feature transformation sequence (e.g., $log(f_1),-f_3,...$) and $v_{i}$ is the corresponding model performance. 
In this method, two agents select candidate features while a third agent selects the transformation operation at each iteration, jointly optimized by maximizing downstream task performance.
For more details, please refer to the referenced paper.

\noindent\textbf{Token-Level Feature Transformation Embedding.}
After collecting feature transformation sequence-accuracy pairs $\{(\mathbf{\Gamma_{i}},v_{i})\}^{n}_{i=1}$, we train the token-level encoder and decoder to capture token-level transformation patterns from feature-operation token sequences.

\noindent\textbf{Encoder $\phi_{tok}$:}
The token encoder aims to learn a mapping function $\phi_{tok}$ that converts the input transformation sequence $\mathbf{\Gamma} \in \mathbb{R}^{1\times N}$ to token embeddings $\mathbf{E}$, denoted by $\mathbf{E} = \phi_{tok}(\mathbf{\Gamma}) \in \mathbb{R}^{N\times d_{tok}}$, where $N$ is the length of the input transformation sequence $\mathbf{\Gamma}$ and $d_{tok}$ is the hidden size of token embedding.
In {\model}, we adopt two Transformer encoder layers~\citep{transformer} as the token encoder. 

\noindent\textbf{Decoder $\psi_{tok}$:}
The token decoder is designed to reconstruct the transformation sequence $\mathbf{\Gamma}$ from the learned token embedding $\mathbf{E}$, which can be denoted by $\mathbf{\Gamma} = \psi_{tok}(\mathbf{E})$.
 In {\model}, we adopt two Transformer decoder layers~\citep{transformer} as the token decoder.

\noindent\textbf{Why Ensuring Permutation Invariance in Concept-Level Embeddings Matters.}
While token-level encoders capture low-level feature--operation interactions, they fail to model high-level transformation patterns.
In practice, a set of generated transformation concepts is semantically invariant to the ordering of its concepts.
However, existing methods encode these concepts in an order-sensitive manner, introducing permutation bias that fragments equivalent transformations across multiple embeddings.
This redundancy distorts the embedding space and inflates the search space, leading to inefficient exploration.
We thus propose a permutation-invariant concept encoder--decoder that maps equivalent concepts to canonical embeddings, enabling faithful representation learning and efficient exploration of the learned embedding space.

\noindent\textbf{Encoder $\phi_{con}$:}
The concept encoder aims to encode the concepts into global embeddings. 
To ensure permutation invariance, we adopt a self-attention pooling mechanism~\citep{set_tf} in $\phi_{con}$.
This structure is inherently permutation-invariant with respect to the ordering of concepts, ensuring that the same concept set yields an identical global embedding regardless of its order.
Specifically, each transformation sequence is represented in postfix form, and each concept is constructed by sequentially composing feature and operation tokens according to this representation.
For simple transformations, we obtain the concept embedding by averaging the embeddings of the involved tokens.
For nested transformations, this aggregation is applied recursively following the postfix order.
For example, $\log(\log(f_7))$ is represented as $(f_7, \log, \log)$ and encoded as
$\text{Mean}(\text{Mean}(\mathbf{E}_{f_7}, \mathbf{E}_{\log}), \mathbf{E}_{\log})$.
Thus, token embeddings $\mathbf{E}$ are aggregated into concept embeddings $\mathbf{G}$ according to the compositional structure of the generated transformations.
% Specifically, we first adopt a mean pooling layer to aggregate the token embeddings $\mathbf{E}$ into concept embeddings $\mathbf{G}$ according to the relationship between concept and its affiliated feature indices and mathematical operations, serving as the aggregation layer.
% For instance, suppose concepts are derived from the transformation $(f_1 + f_7, log(f_2),...)$, then corresponding concept embeddings are obtained by averaging the token embeddings of the involved features and mathematical operations, denoted by $\mathbf{G}= (\mathbf{G}_1,\mathbf{G}_2,...)$, where $\mathbf{G}_1=\text{Mean}(\textbf{E}_{f_1},\textbf{E}_{plus},\textbf{E}_{f_7})$ and $\mathbf{G}_2 = \text{Mean}(\textbf{E}_{f_2},\textbf{E}_{log})$.
Then, we initialize a set of $k$ learnable seed vectors $\mathbf{\mathcal{S}} \in \mathbb{R}^{k \times d_{seed}}$ using Xavier uniform initialization, where $k$ is the number of seed vectors and $d_{seed}$ is the hidden size of each seed vector.
These learnable seed vectors serve as queries in the multi-head attention module, attending over the concept embeddings $\mathbf{G}$ which serve as keys and values to extract high-level semantic concept representations.
The output is then combined with the original seed vectors through a residual connection, followed by a linear layer. 
This yields the global embedding, denoted by: $\mathbf{G}' = \phi_{con}(\mathbf{G})= \text{Linear}(\mathbf{\mathcal{S}} + \text{Multihead}(\mathbf{\mathcal{S}}, \mathbf{G}, \mathbf{G})) \in \mathbb{R}^{k\times d_{global}}$, where $d_{global}$ represents the hidden size of global embedding.

\noindent\textbf{Decoder $\psi_{con}$:}
The concept decoder $\psi_{con}$ aims to reconstruct concept embedding $\mathbf{G}$ from the global embedding $\mathbf{G}'$, denoted by $\mathbf{G} = \psi_{con}(\mathbf{G}')$. 
In {\model}, we adopt the same architecture for the concept decoder as used in the concept encoder, ensuring symmetric abstraction and reconstruction. 

\begin{algorithm}[t]
\caption{Training and Search Procedure of {\model}}
\label{alg:pher}
\KwIn{Transformation records $\{(\mathbf{\Gamma}_i, v_i)\}_{i=1}^{n}$, top-$K$ seeds, search epochs $T$}
\KwOut{Optimal transformation sequence $\mathbf{\Gamma}^{*}$}
\tcp{Stage 1: Token-Level Training}
\For{each epoch}{
    $\mathbf{E} \leftarrow \phi_{tok}(\mathbf{\Gamma})$, $\hat{\mathbf{\Gamma}} \leftarrow \psi_{tok}(\mathbf{E})$\;
    Update $\phi_{tok}, \psi_{tok}$ by minimizing $\mathcal{L}_{tok}$\;
}
\tcp{Stage 2: Concept-Level Training}
Freeze $\phi_{tok}$\;
\For{each epoch}{
    $\mathbf{E} \leftarrow \phi_{tok}(\mathbf{\Gamma})$, $\mathbf{G} \leftarrow \text{Aggregate}(\mathbf{E})$\;
    $\mathbf{G}' \leftarrow \phi_{con}(\mathbf{G})$, $\hat{\mathbf{G}} \leftarrow \psi_{con}(\mathbf{G}')$\;
    Update $\phi_{con}, \psi_{con}$ by minimizing $\mathcal{L}_{con}$\;
}
\tcp{Stage 3: Token-Concept Alignment}
Freeze $\phi_{con}^{*}, \psi_{con}^{*}$\;
\For{each epoch}{
    $\hat{\mathbf{\Gamma}} \leftarrow \psi_{tok}(\psi_{con}^{*}(\phi_{con}^{*}(\text{Aggregate}(\phi_{tok}(\mathbf{\Gamma})))))$\;
    Update $\phi_{tok}, \psi_{tok}$ by minimizing $\mathcal{L}_{align}$\;
}
\tcp{Stage 4: Policy-Guided Search}
Select top-$K$ records by performance as search seeds\;
Initialize PPO agent $\mathcal{A}$ with actor-critic architecture\;
\For{$t = 1$ \KwTo $T$}{
    \For{each seed $(\mathbf{\Gamma}, v)$}{
        $\mathbf{G}' \leftarrow \phi_{con}^{*}(\text{Aggregate}(\phi_{tok}^{*}(\mathbf{\Gamma})))$\;
        $\mathbf{G}'_{+} \leftarrow \mathcal{A}(\mathbf{G}')$\;
        $\mathbf{\Gamma}^{+} \leftarrow \psi_{tok}^{*}(\psi_{con}^{*}(\mathbf{G}'_{+}))$\;
        Compute reward $\mathcal{R}$ based on $\mathcal{M}(X[\mathbf{\Gamma}^{+}])$ and $|\mathbf{\Gamma}^{+}|$\;
    }
    Update $\mathcal{A}$ by maximizing $\mathcal{L}_{actor}$ and minimizing $\mathcal{L}_{critic}$\;
}
$\mathbf{\Gamma}^{*} \leftarrow \arg\max_{\mathbf{\Gamma}^{+}} \mathcal{M}(X[\mathbf{\Gamma}^{+}])$\;
\Return $\mathbf{\Gamma}^{*}$\;
\end{algorithm}

\noindent\textbf{Hierarchical Optimization Procedure:}
To effectively train the hierarchical encoder-decoder models, we design a three-stage training strategy.
Each stage progressively captures feature transformation knowledge, from token-level to concept-level.
% For notational simplicity, the aggregation layer is omitted in the loss functions.

\noindent\textbf{Stage 1: Token-Level Training.}
In the first stage, we train the token encoder $\phi_{tok}$ and decoder $\psi_{tok}$ to reconstruct the input transformation sequence $\mathbf{\Gamma}$.
Specifically, given a sequence $\mathbf{\Gamma}$, the token encoder first produces token embedding $\mathbf{E} = \phi_{tok}(\mathbf{\Gamma})$.
The token decoder then reconstructs the original feature transformation sequence from the token embedding $\mathbf{\hat{\Gamma}} = \psi_{tok}(\mathbf{E})$.
We optimize the token-level model by minimizing the negative log-likelihood loss:
\vspace{-0.1cm}
\begin{equation}
    \mathcal{L}_{tok} = -\log P_{\psi_{tok}}(\mathbf{\Gamma} \mid \mathbf{E}).
\end{equation}

\noindent\textbf{Stage 2: Concept-Level Training.}
After training the token-level model, we freeze its parameters and train the concept encoder $\phi_{con}$ and decoder $\psi_{con}$ to learn high-level permutation-invariant global embeddings. 
Specifically, given a feature transformation sequence $\mathbf{\Gamma}$, we first utilize the well-trained token encoder to obtain token embedding $\mathbf{E}$.
Then, we input $\mathbf{E}$ into the aggregation layer to generate the concept embedding $\mathbf{G}$ based on the relationships between concept and its affiliated feature indices and mathematical operations.
Thereafter, $\mathbf{G}$ is input into $\phi_{con}$ to obtain the global embedding 
$\mathbf{G}' = \phi_{con}(\mathbf{G})$. 
Finally, the global embedding $\mathbf{G}'$ is decoded back to concept embedding $\hat{\mathbf{G}} = \psi_{con}(\mathbf{G}')$.
We optimize the concept-level model by minimizing the Mean Squared Error (MSE) between the reconstructed $\hat{\mathbf{G}}$ and original concept embedding $\mathbf{G}$, denoted by:
% \vspace{-0.1cm}
\begin{equation}
    \mathcal{L}_{con} = \text{MSE}(\mathbf{G},\hat{\mathbf{G}})= \text{MSE}(\mathbf{G},\psi_{con}(\phi_{con}(\mathbf{G}))).
\end{equation}

\noindent\textbf{Stage 3: Token-Concept Alignment.}
In this stage, we freeze the concept-level model and tune the token encoder and decoder to improve the accuracy of end-to-end token reconstruction.
Given a transformation sequence $\mathbf{\Gamma}$, we obtain the token embedding $\mathbf{E} = \phi_{tok}(\mathbf{\Gamma})$.
% Then, $\mathbf{E}$ is input into the well-trained concept encoder $\phi_{con}^{*}$, the well-trained concept decoder $\psi_{con}^{*}$ and the token decoder in sequence to reconstruct the feature transformation sequence.
Then, the aggregation layer maps $\mathbf{E}$ to concept embeddings $\mathbf{G}$, which are further processed by the well-trained concept encoder $\phi_{con}^{*}$ and concept decoder $\psi_{con}^{*}$ before being decoded by the token decoder.
We optimize the following loss during the alignment stage:
$
\mathcal{L}_{align} = -\log P_{\psi_{tok}}(\mathbf{\Gamma} \mid \psi_{\text{con}}^{*}(\phi_{\text{con}}^{*}(\phi_{tok}(\mathbf{\Gamma})))).
$
Here, the aggregation layer is omitted for brevity.
This stage aligns token embeddings with the fixed concept-level abstractions, improving the coherence and semantic quality of the reconstructed transformation sequence.
The complete training and search procedure of {\model} is summarized in Algorithm~\ref{alg:pher}.

% \vspace{0.2cm}
\subsection{Policy-guided Multi-objective Search}
\noindent\textbf{Why selecting search seeds matters.}
Once the global embedding space has converged, we employ a policy-guided multi-objective search strategy in the learned global embedding space to identify the optimal global embedding that achieves maximum downstream task performance and minimum transformation length.
Inspired by the significance of initialization for deep neural networks, good starting points are crucial for accelerating the search process and enhancing its performance.
Thus, we rank all collected records by model performance and select the top-K as search seeds.

\noindent\textbf{Policy-guided multi-objective search.}
Existing methods assume the learned embedding space is convex, relying on gradient-ascent search strategy to identify the global optimal embedding.
However, due to the complex interactions between features and mathematical operations, the embedding space is highly non-convex in practice, increasing the risk of becoming trapped in local optima and resulting in suboptimal feature transformation sequences.
Thus, to perform effective exploration and search process without strong convexity assumptions, we employ Proximal Policy Optimization (PPO)~\citep{ppo} method, which is a widely used policy gradient algorithm for multi-objective optimization.
PPO stabilizes policy updates by clipping and trust region mechanisms, effectively balancing exploration and exploitation in high-dimensional and non-convex embedding spaces.
In {\model}, the PPO agent learns to explore the continuous global embedding space by selecting promising global embeddings based on the downstream task performance and transformation sequence length.
Specifically, given a global embedding $\mathbf{G}'$, the PPO agent $\mathcal{A}$ aims to manipulate it to generate the optimal global embedding $\mathbf{G}^{'}_{opt}$, denoted by:
$
    \mathcal{A}(\mathbf{G}^{'}) = \mathbf{G}^{'}_{opt}.
$

\noindent\textbf{Reward $\mathcal{R}$:}
To balance the performance and the total length of the transformed sequence, we propose a weighted reward function, denoted by:
$
    \mathcal{R} = \lambda(\mathcal{M}(X[\mathbf{\Gamma^{+}}]) - \mathcal{M}(X[\mathbf{\Gamma}])) + (1 - \lambda) \mathcal{N}[\mathbf{\Gamma^{+}}],
$
where $\mathcal{M}$ denotes the downstream ML task, 
$\mathbf{\Gamma}$ represents the original transformation sequence, 
$\mathbf{\Gamma^{+}}$ refers to the searched transformation sequence, 
$\mathcal{N}[\mathbf{\Gamma^{+}}]$ is a normalized length penalty of $\mathbf{\Gamma^{+}}$, 
and $\lambda$ is the trade-off hyperparameter to balance the performance improvement and the total length of the transformed sequence.

\noindent\textbf{Solving the Optimization Problem.}
We adopt a multi-objective optimization strategy to train the PPO agent.
Specifically, the policy is updated by maximizing a clipped surrogate objective, which approximates the cumulative discounted reward obtained throughout the iterative search process, while constraining the step size of each update.
To enable the PPO agent to identify more informative embeddings with higher model performance and fewer transformation operations, we adopt an actor-critic model architecture.
We optimize the critic by minimizing the difference between the observed cumulative discounted rewards and their predictions, denoted by:
$
    \mathcal{L}_{critic} = \frac{1}{T}\sum^{T}_{t=1}(V(\mathbf{s}_t)-G_t)^{2} 
    ,
$
where $T$ is the trajectory length, $\mathbf{s}_t$ is the state at time step $t$, $V(\mathbf{s}_t)$ is the predicted value from the critic,  $G_t$ is the cumulative discounted return, and $\gamma\in [0,1]$ is the discounted factor.
The actor is optimized by maximizing a clipped surrogate objective while constraining the step size of each policy update, denoted by:
$
    \mathcal{L}_{actor} = \hat{\mathbb{E}}_{t}\left[ \min(r_t(\theta)\hat{A}_t, \text{clip}(r_t(\theta), 1-\epsilon,1+\epsilon)\hat{A}_t)\right],
$
where $\epsilon$ is the clipping ratio hyperparameter, 
$r_t(\theta)
$ 
is the probability ratio, and $\hat{A}_t 
$ is the estimated advantage.
Once the actor network converges, the learned policy $\pi^*$ guides the global embedding $\mathbf{G}^{'}$ within the global embedding space $\mathbb{G}$ toward regions that yield higher downstream task performance with fewer transformation operations.
Then, we reconstruct the transformation sequence $\mathbf{\Gamma}^{+}$ from the enhanced global embedding $\mathbf{G}^{'}_{+}$ using the well-trained concept decoder and token decoder.
Next, we obtain enhanced feature spaces using the enhanced transformation sequences $[\mathbf{\Gamma}^{+}_1, \mathbf{\Gamma}^{+}_2,..., \mathbf{\Gamma}^{+}_K]$ and input them into the downstream ML task to evaluate their performance.
Finally, the feature transformation sequence achieving the highest performance is selected as the optimal sequence, denoted as $\mathbf{\Gamma}^{*}$. 
% \vspace{-0.9cm}

\begin{table*}[!t]
    \centering
    % \vspace{-0.7cm}
    \caption{Overall performance comparison. In this table, the best and second-best results are highlighted in \textcolor{red}{\textbf{red}} and \textcolor{blue}{blue}, respectively. (\textbf{The higher the value is, the better the model performance is.})}
    \vspace{-0.2cm}
    \label{main_table}
    \resizebox{\linewidth}{!}{
        \begin{tabular}{ccccccccccccc}
            \toprule
            Dataset            & C/R & Original & RDG                & ERG                                  & LDA                & AFAT               & NFS                & TTG                & GRFG                                 & DIFER              & MOAT                                 & {\model}                                     \\               \midrule
            Higgs Boson        & C   & 0.696    & 0.700$^{\pm0.001}$ & $\textcolor{blue}{0.704}^{\pm0.003}$ & 0.511$^{\pm0.011}$ & 0.695$^{\pm0.001}$ & 0.697$^{\pm0.002}$ & 0.699$^{\pm0.003}$ & 0.701$^{\pm0.002}$                   & 0.669$^{\pm0.001}$ & 0.699$^{\pm0.002}$                   & \textbf{\textcolor{red}{0.706}}$^{\pm0.002}$ \\
            Amazon Employee    & C   & 0.930    & 0.931$^{\pm0.001}$ & $\textcolor{blue}{0.935}^{\pm0.001}$ & 0.914$^{\pm0.002}$ & 0.933$^{\pm0.002}$ & 0.932$^{\pm0.001}$ & 0.930$^{\pm0.001}$ & 0.932$^{\pm0.001}$                   & 0.929$^{\pm0.001}$ & 0.933$^{\pm0.002}$                   & \textbf{\textcolor{red}{0.936}}$^{\pm0.002}$ \\
            PimaIndian         & C   & 0.776    & 0.760$^{\pm0.007}$ & 0.762$^{\pm0.004}$                   & 0.729$^{\pm0.061}$ & 0.760$^{\pm0.011}$ & 0.759$^{\pm0.014}$ & 0.750$^{\pm0.019}$ & 0.754$^{\pm0.011}$                   & 0.760$^{\pm0.013}$ & $\textcolor{blue}{0.807}^{\pm0.011}$ & \textbf{\textcolor{red}{0.828}}$^{\pm0.003}$ \\
            SpectF             & C   & 0.760    & 0.760$^{\pm0.001}$ & 0.759$^{\pm0.018}$                   & 0.665$^{\pm0.119}$ & 0.782$^{\pm0.079}$ & 0.760$^{\pm0.001}$ & 0.775$^{\pm0.014}$ & 0.818$^{\pm0.001}$                   & 0.766$^{\pm0.002}$ & $\textcolor{blue}{0.912}^{\pm0.014}$ & \textbf{\textcolor{red}{0.929}}$^{\pm0.003}$ \\
            SVMGuide3          & C   & 0.778    & 0.791$^{\pm0.008}$ & 0.817$^{\pm0.015}$                   & 0.635$^{\pm0.049}$ & 0.789$^{\pm0.009}$ & 0.786$^{\pm0.004}$ & 0.791$^{\pm0.014}$ & 0.812$^{\pm0.025}$                   & 0.773$^{\pm0.021}$ & $\textcolor{blue}{0.845}^{\pm0.001}$ & \textbf{\textcolor{red}{0.848}}$^{\pm0.003}$ \\
            German Credit      & C   & 0.649    & 0.667$^{\pm0.008}$ & 0.684$^{\pm0.011}$                   & 0.597$^{\pm0.058}$ & 0.640$^{\pm0.032}$ & 0.663$^{\pm0.017}$ & 0.663$^{\pm0.019}$ & 0.683$^{\pm0.013}$                   & 0.656$^{\pm0.018}$ & $\textcolor{blue}{0.729}^{\pm0.001}$ & \textbf{\textcolor{red}{0.736}}$^{\pm0.007}$ \\
            Credit Default     & C   & 0.802    & 0.805$^{\pm0.001}$ & 0.804$^{\pm0.002}$                   & 0.743$^{\pm0.009}$ & 0.804$^{\pm0.001}$ & 0.803$^{\pm0.002}$ & 0.803$^{\pm0.002}$ & 0.806$^{\pm0.003}$                   & 0.796$^{\pm0.005}$ & $\textcolor{blue}{0.808}^{\pm0.001}$ & \textbf{\textcolor{red}{0.810}}$^{\pm0.001}$ \\
            Messidor\_features & C   & 0.633    & 0.685$^{\pm0.053}$ & 0.665$^{\pm0.025}$                   & 0.463$^{\pm0.084}$ & 0.656$^{\pm0.005}$ & 0.657$^{\pm0.011}$ & 0.662$^{\pm0.019}$ & 0.692$^{\pm0.033}$                   & 0.660$^{\pm0.007}$ & $\textcolor{blue}{0.694}^{\pm0.005}$ & \textbf{\textcolor{red}{0.744}}$^{\pm0.003}$ \\
            Wine Quality Red   & C   & 0.456    & 0.514$^{\pm0.016}$ & 0.494$^{\pm0.005}$                   & 0.401$^{\pm0.061}$ & 0.480$^{\pm0.042}$ & 0.467$^{\pm0.021}$ & 0.464$^{\pm0.010}$ & 0.470$^{\pm0.009}$                   & 0.476$^{\pm0.018}$ & $\textcolor{blue}{0.559}^{\pm0.002}$ & \textbf{\textcolor{red}{0.562}}$^{\pm0.003}$ \\
            Wine Quality White & C   & 0.536    & 0.524$^{\pm0.001}$ & 0.524$^{\pm0.001}$                   & 0.437$^{\pm0.015}$ & 0.516$^{\pm0.032}$ & 0.533$^{\pm0.010}$ & 0.529$^{\pm0.002}$ & 0.534$^{\pm0.019}$                   & 0.507$^{\pm0.025}$ & $\textcolor{blue}{0.536}^{\pm0.010}$ & \textbf{\textcolor{red}{0.553}}$^{\pm0.002}$ \\
            SpamBase           & C   & 0.927    & 0.924$^{\pm0.001}$ & 0.920$^{\pm0.001}$                   & 0.885$^{\pm0.030}$ & 0.917$^{\pm0.001}$ & 0.922$^{\pm0.003}$ & 0.922$^{\pm0.003}$ & 0.922$^{\pm0.005}$                   & 0.912$^{\pm0.016}$ & $\textcolor{blue}{0.932}^{\pm0.001}$ & \textbf{\textcolor{red}{0.936}}$^{\pm0.002}$ \\
            AP-omentum-ovary   & C   & 0.845    & 0.845$^{\pm0.001}$ & 0.814$^{\pm0.001}$                   & 0.710$^{\pm0.134}$ & 0.845$^{\pm0.125}$ & 0.845$^{\pm0.001}$ & 0.845$^{\pm0.001}$ & $\textcolor{blue}{0.849}^{\pm0.001}$ & 0.833$^{\pm0.031}$ & 0.845$^{\pm0.002}$                   & \textbf{\textcolor{red}{0.883}}$^{\pm0.001}$ \\
            Lymphography       & C   & 0.141    & 0.108$^{\pm0.001}$ & 0.129$^{\pm0.009}$                   & 0.144$^{\pm0.132}$ & 0.150$^{\pm0.133}$ & 0.170$^{\pm0.035}$ & 0.174$^{\pm0.040}$ & 0.182$^{\pm0.013}$                   & 0.150$^{\pm0.116}$ & $\textcolor{blue}{0.267}^{\pm0.035}$ & \textbf{\textcolor{red}{0.410}}$^{\pm0.105}$ \\
            MNIST fashion      & C   & 0.712    & 0.714$^{\pm0.001}$ & 0.716$^{\pm0.001}$                   & 0.510$^{\pm0.020}$ & 0.699$^{\pm0.022}$ & 0.710$^{\pm0.002}$ & 0.711$^{\pm0.004}$ & 0.728$^{\pm0.001}$                   & 0.717$^{\pm0.002}$ & $\textcolor{blue}{0.729}^{\pm0.001}$ & \textbf{\textcolor{red}{0.735}}$^{\pm0.001}$ \\
            Housing Boston     & R   & 0.415    & 0.420$^{\pm0.031}$ & 0.418$^{\pm0.001}$                   & 0.021$^{\pm0.002}$ & 0.423$^{\pm0.005}$ & 0.424$^{\pm0.002}$ & 0.421$^{\pm0.011}$ & 0.404$^{\pm0.007}$                   & 0.381$^{\pm0.017}$ & $\textcolor{blue}{0.465}^{\pm0.006}$ & \textbf{\textcolor{red}{0.538}}$^{\pm0.005}$ \\
            Airfoil            & R   & 0.519    & 0.520$^{\pm0.001}$ & 0.519$^{\pm0.001}$                   & 0.207$^{\pm0.065}$ & 0.509$^{\pm0.001}$ & 0.519$^{\pm0.002}$ & 0.521$^{\pm0.001}$ & 0.521$^{\pm0.003}$                   & 0.558$^{\pm0.002}$ & $\textcolor{blue}{0.627}^{\pm0.011}$ & \textbf{\textcolor{red}{0.645}}$^{\pm0.001}$ \\
            Openml\_589        & R   & 0.510    & 0.548$^{\pm0.032}$ & 0.610$^{\pm0.001}$                   & 0.034$^{\pm0.079}$ & 0.509$^{\pm0.002}$ & 0.506$^{\pm0.004}$ & 0.502$^{\pm0.002}$ & 0.627$^{\pm0.006}$                   & 0.463$^{\pm0.005}$ & $\textcolor{blue}{0.656}^{\pm0.001}$ & \textbf{\textcolor{red}{0.660}}$^{\pm0.006}$ \\
            Openml\_618        & R   & 0.469    & 0.444$^{\pm0.026}$ & 0.543$^{\pm0.039}$                   & 0.030$^{\pm0.076}$ & 0.473$^{\pm0.002}$ & 0.471$^{\pm0.004}$ & 0.472$^{\pm0.001}$ & 0.562$^{\pm0.101}$                   & 0.408$^{\pm0.036}$ & $\textcolor{blue}{0.692}^{\pm0.002}$ & \textbf{\textcolor{red}{0.693}}$^{\pm0.001}$ \\
            Openml\_620        & R   & 0.510    & 0.494$^{\pm0.019}$ & 0.541$^{\pm0.008}$                   & 0.026$^{\pm0.013}$ & 0.520$^{\pm0.001}$ & 0.509$^{\pm0.005}$ & 0.513$^{\pm0.004}$ & 0.568$^{\pm0.013}$                   & 0.442$^{\pm0.004}$ & $\textcolor{blue}{0.643}^{\pm0.003}$ & \textbf{\textcolor{red}{0.652}}$^{\pm0.003}$ \\ \bottomrule
        \end{tabular}}
    \begin{tablenotes}
        \small
        \item * We evaluated classification (C) and regression (R) tasks in terms of F1-Score and 1-RAE, respectively.
        \item * The standard deviation is computed based on the results of 5 independent runs.
    \end{tablenotes}
    % \vspace{-0.6cm}
\end{table*}

\begin{figure*}[!t]
    % \vspace{-0.3cm}
    \centering
    \subfigure[Spectf]{
        \includegraphics[width=4.25cm, trim={0 0 0.5cm 0}]{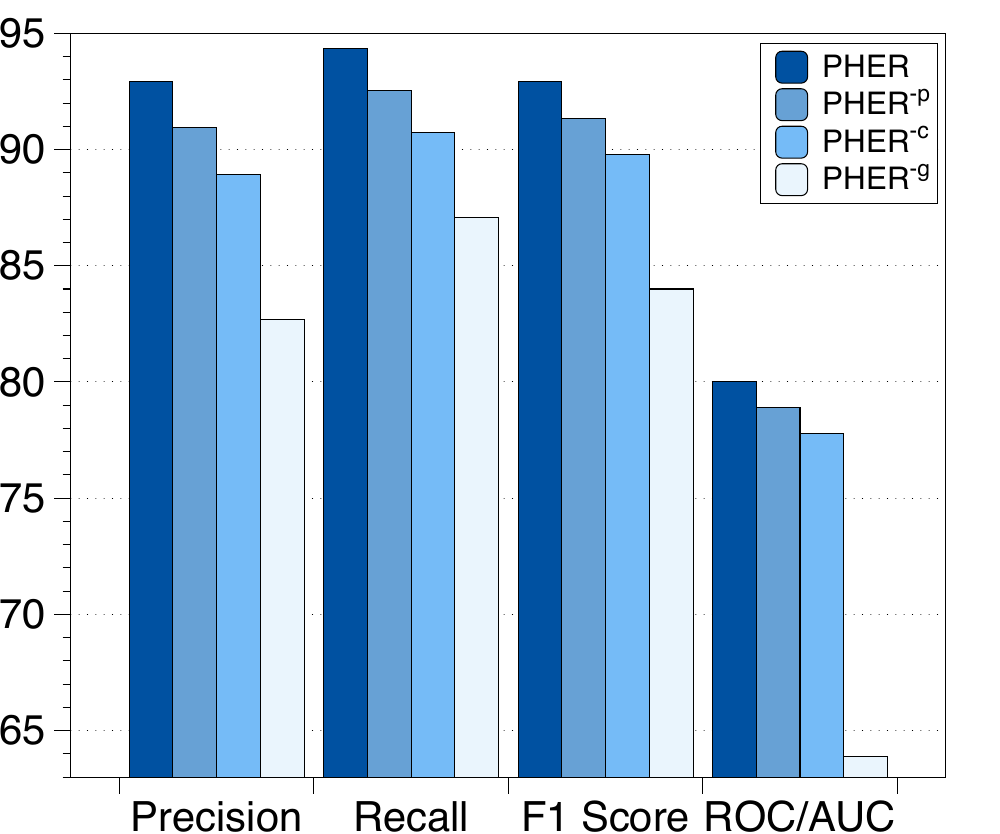}
    }
    \hspace{-3mm}
    \subfigure[PimaIndian]{
        \includegraphics[width=4.25cm, trim={0 0 0.5cm 0}]{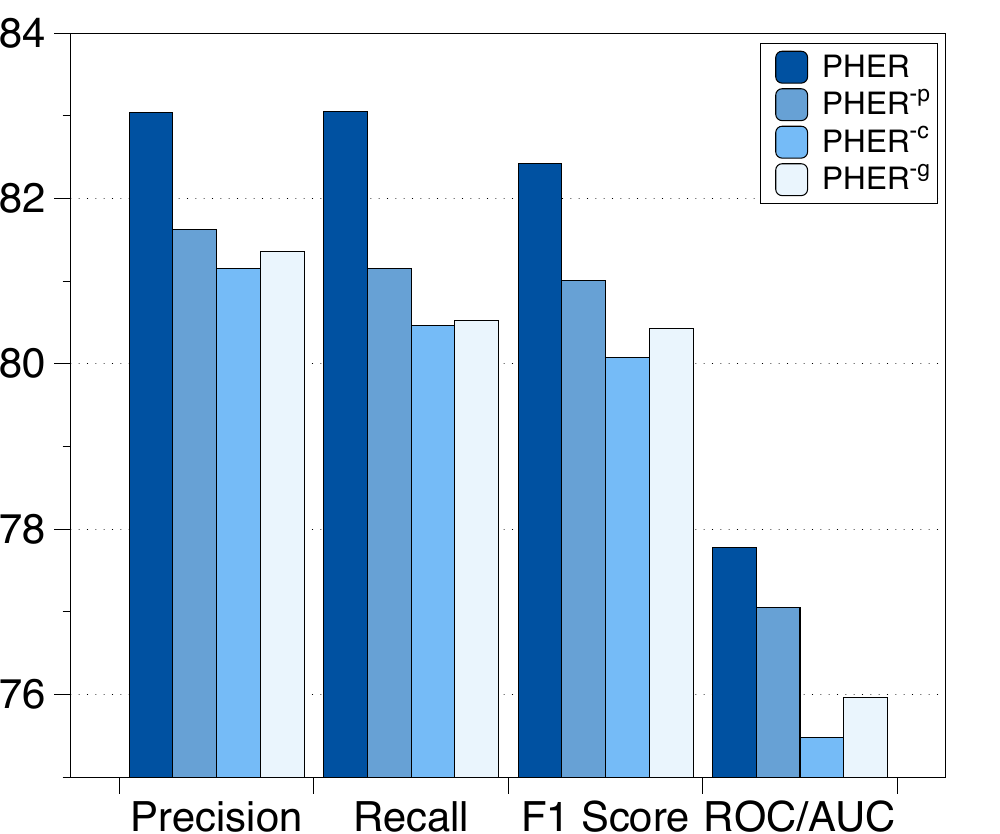}
    }
    \hspace{-3mm}
    \subfigure[Openml\_589]{
        \includegraphics[width=4.25cm, trim={0 0 0.5cm 0}]{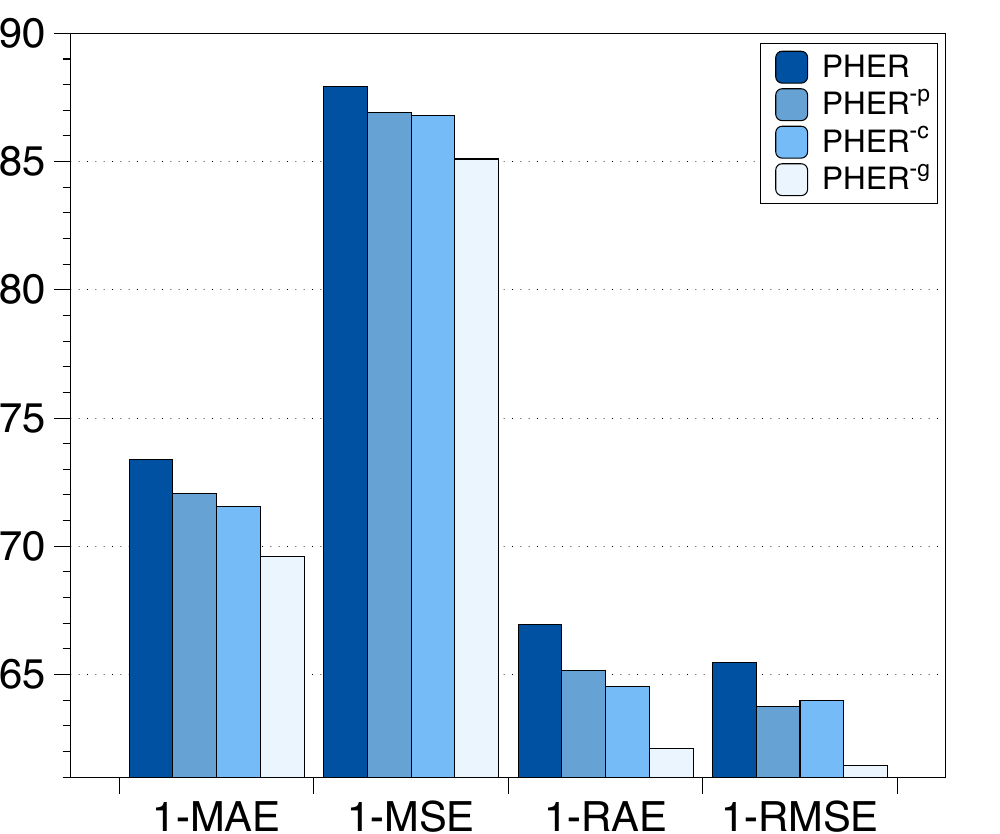}
    }
    \hspace{-3mm}
    \subfigure[Openml\_620]{
        \includegraphics[width=4.25cm, trim={0 0 0.5cm 0}]{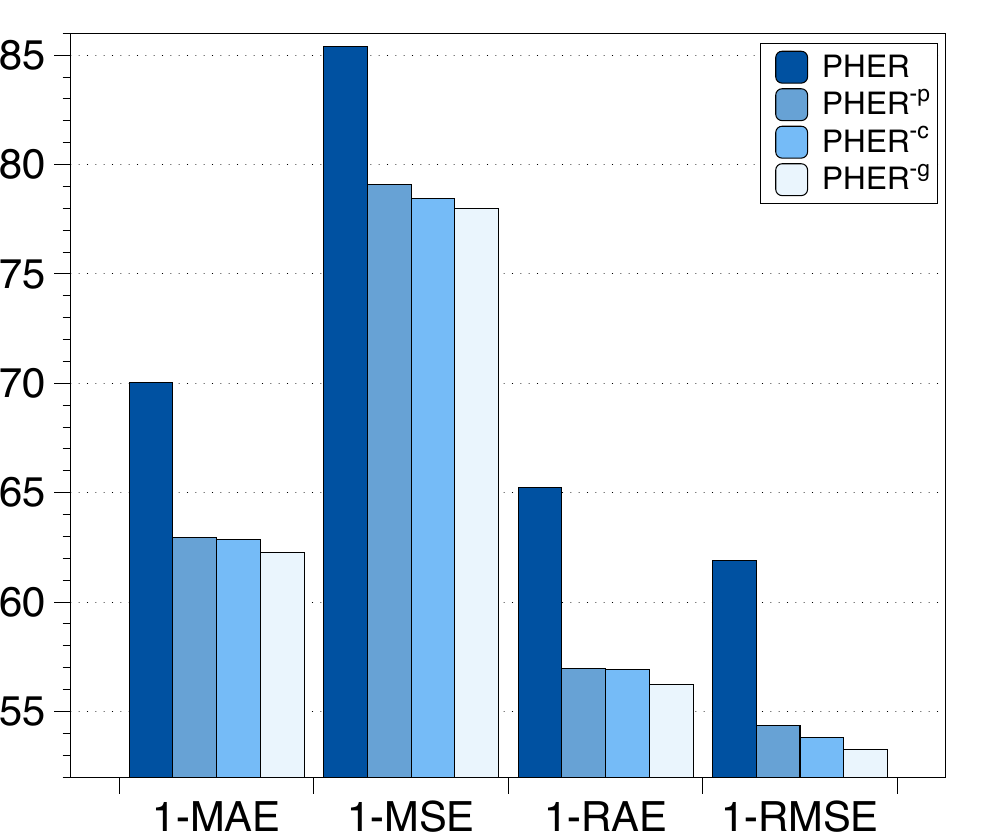}
    }
    \vspace{-0.35cm}
    \caption{The influence of hierarchical modeling (\model$^{-c}$), permutation invariance (\model$^{-p}$), and policy-guided search (\model$^{-g}$) in \model.}
    \label{ablation}
    % \vspace{-0.25cm}
\end{figure*}

\vspace{-0.3cm}
\section{Experiments}
\label{experiment_section}
% \subsection{Experimental Platform Information}
% \label{appendix:experimental_platform}
% We conduct all the experiments on the Windows 11 operating system with an AMD Ryzen~5~5600X CPU and an NVIDIA GeForce RTX~3070Ti GPU, using Python 3.10.15 and PyTorch 2.5.1.
% To enhance the reproducibility of our framework, we further provide hyperparameter settings in Appendix~\ref{appendix:hyperparams}.

% \vspace{-0.3cm}
\subsection{Datasets and Evaluation Metrics}
We evaluate {\model} on 19 publicly accessible datasets from UCI~\cite{uci}, LibSVM~\cite{libsvm}, Kaggle~\cite{kaggle}, and OpenML~\cite{openml}, including 14 classification tasks and 5 regression tasks.
% Table~\ref{dataset_info} in Appendix~\ref{dataset_stat} summarizes the detailed statistics of these datasets.
% We adopt Random Forest as the unified downstream model and evaluate performance using five-fold cross-validation.
% All experiments follow a hold-out evaluation protocol and are independently repeated five times with different random seeds, with results reported as the mean and standard deviation.
We adopt Random Forest as the unified downstream model and evaluate its performance via five-fold cross-validation on the training partition.
All experiments follow a hold-out evaluation protocol, where each dataset is randomly divided into 80\% for training and 20\% for testing, and are independently repeated five times with different random seeds, with results reported as the mean and standard deviation.
For regression tasks, we report 1-Relative Absolute Error (1-RAE), 1-Mean Absolute Error (1-MAE), 1-Mean Squared Error (1-MSE), and 1-Root Mean Squared Error (1-RMSE).
For classification tasks, we use F1-score, Precision, Recall, and ROC/AUC.

% \vspace{-0.4cm}
\subsection{Baseline Models}
% \vspace{-0.2cm}
We compare our method with nine widely-used feature transformation methods:
(1) RDG generates new feature-operation-feature transformation records by randomly selecting candidate features and operations.
(2) ERG first applies operations to each feature to expand the feature space and then selects informative features as new features.
(3) LDA~\cite{lda} is a generative probabilistic model that reduces dimensionality by learning latent topic representations.
(4) AFAT~\cite{afat} iteratively applies transformations and performs multi-step feature selection to identify informative ones.
(5) NFS~\cite{nfs} uses a recurrent neural network-based controller trained with reinforcement learning to sequentially model and optimize the transformation process for each feature.
(6) TTG~\cite{ttg} formulates feature transformation as a graph exploration problem and employs reinforcement learning to discover optimal transformation paths within the graph.
(7) GRFG~\cite{grfg} employs three reinforced agents with a feature grouping strategy to perform feature generation in a cascaded manner.
(8) DIFER~\cite{difer} encodes randomly generated transformation sequences into a continuous embedding space and employs greedy gradient-based search to identify the best transformed features.
(9) MOAT~\cite{moat} embeds RL-collected transformation sequences into continuous embeddings using postfix expression and conducts gradient-ascent search with the beam search strategy.
Besides, we developed three variants of {\model} to evaluate the impact of each technical component:
(i) $\textbf{\model}^{-c}$ removes the concept encoder and decoder.
(ii) $\textbf{\model}^{-p}$ replaces the permutation-invariant concept encoder and decoder with permutation-sensitive self-attention layers.
(iii) $\textbf{\model}^{-g}$ replaces the RL search with Genetic Algorithm (GA).
% To ensure fair evaluation, we randomly divide each dataset into 80\% for training and 20\% for testing.
% This experimental setting prevents any test data leakage and ensures a more reliable comparison of feature transformation performance.

\begin{figure*}[!t]
    \vspace{-0.8cm}
    \centering
    \subfigure[RandomForest]{
        \includegraphics[width=4.25cm, trim = {0 0 1cm 0}]{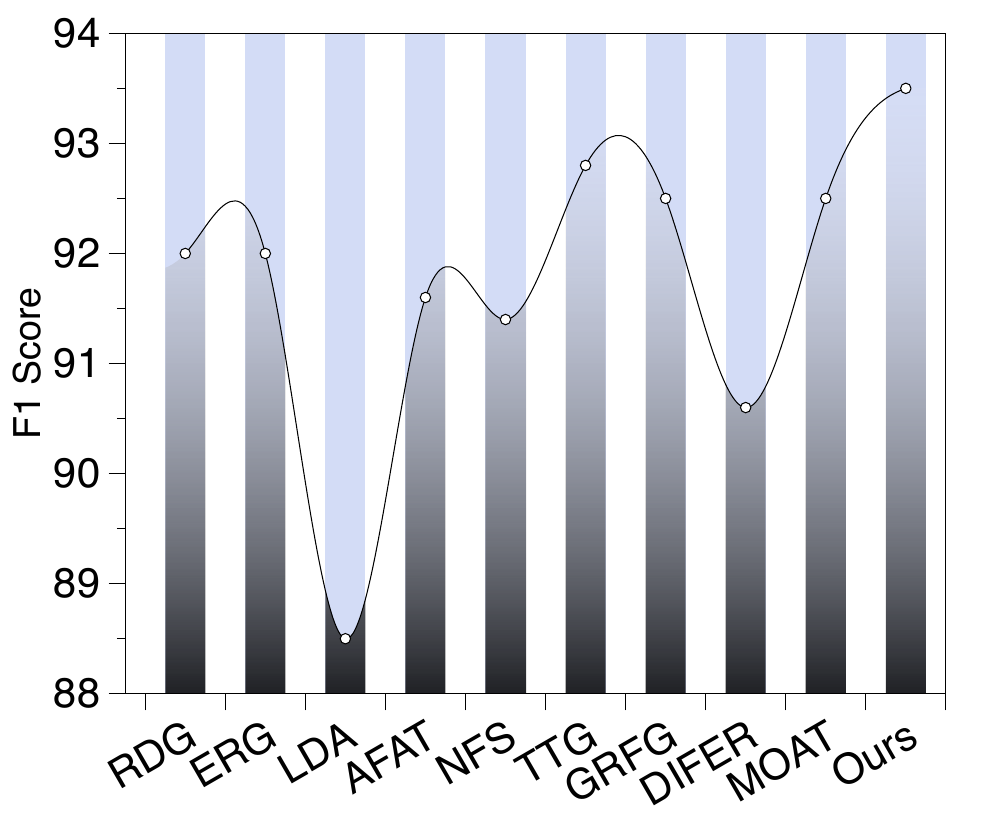}
    }
    \hspace{-3mm}
    \subfigure[XGBoost]{
        \includegraphics[width=4.25cm, trim = {0 0 1cm 0}]{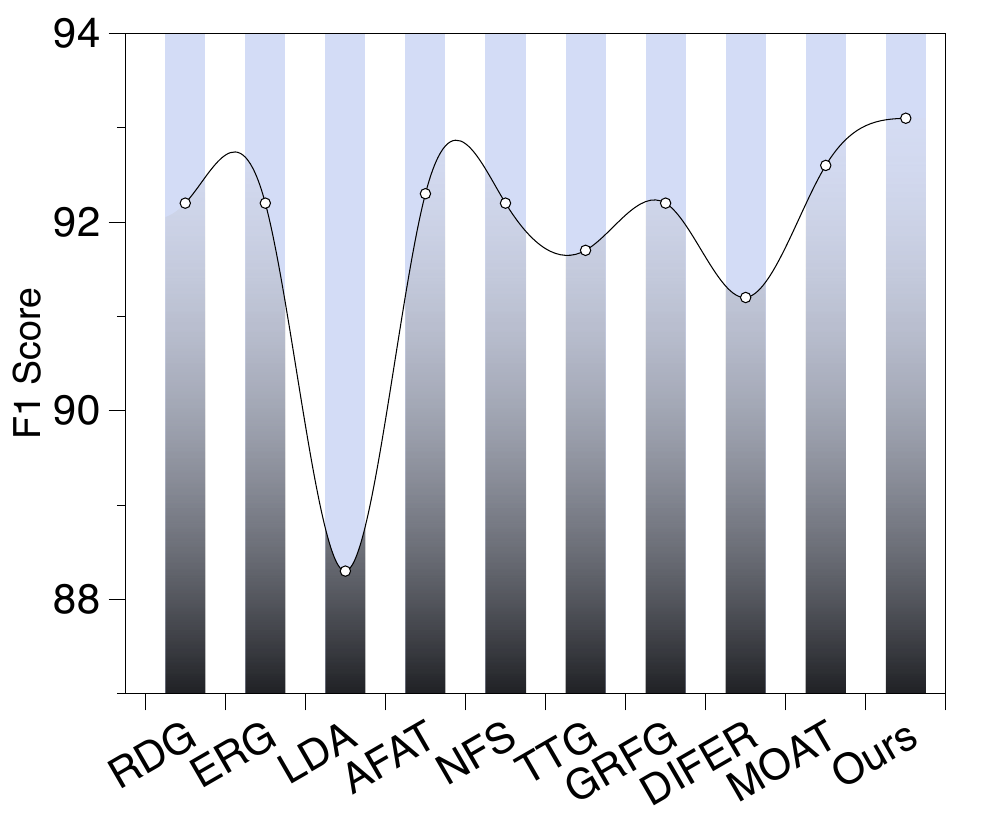}
    }
    \hspace{-3mm}
    \subfigure[KNN]{
        \includegraphics[width=4.25cm, trim = {0 0 1cm 0}]{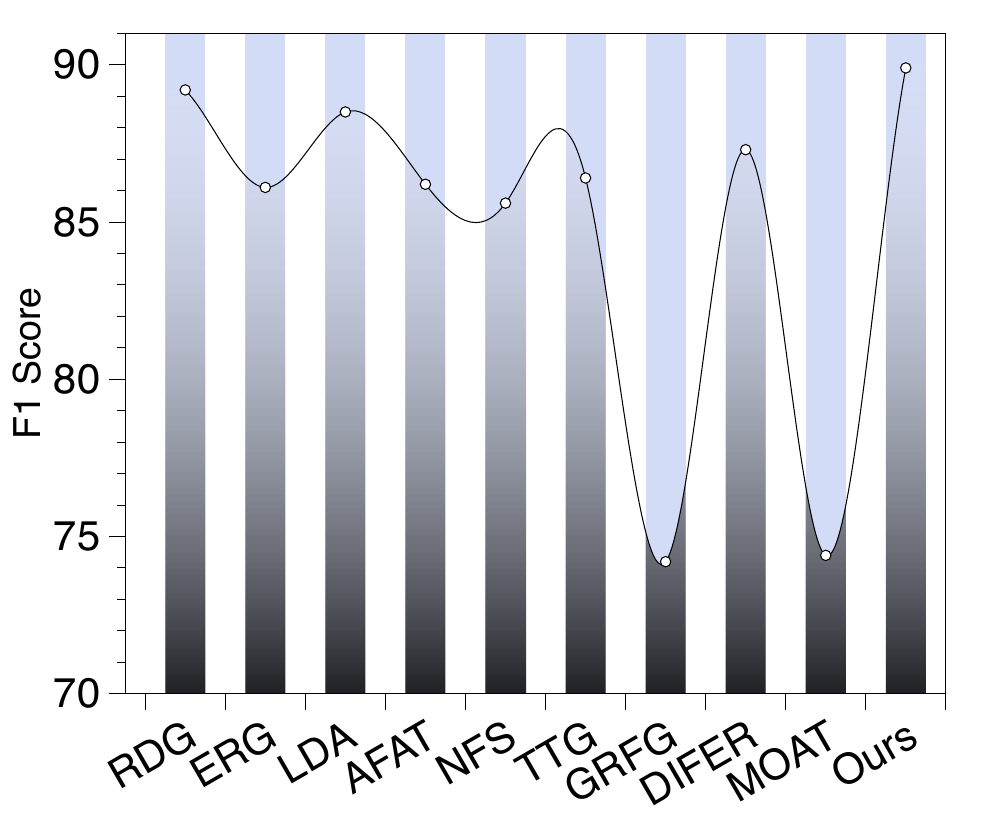}
    }
    \hspace{-3mm}
    \subfigure[Decision Tree]{
        \includegraphics[width=4.25cm, trim = {0 0 1cm 0}]{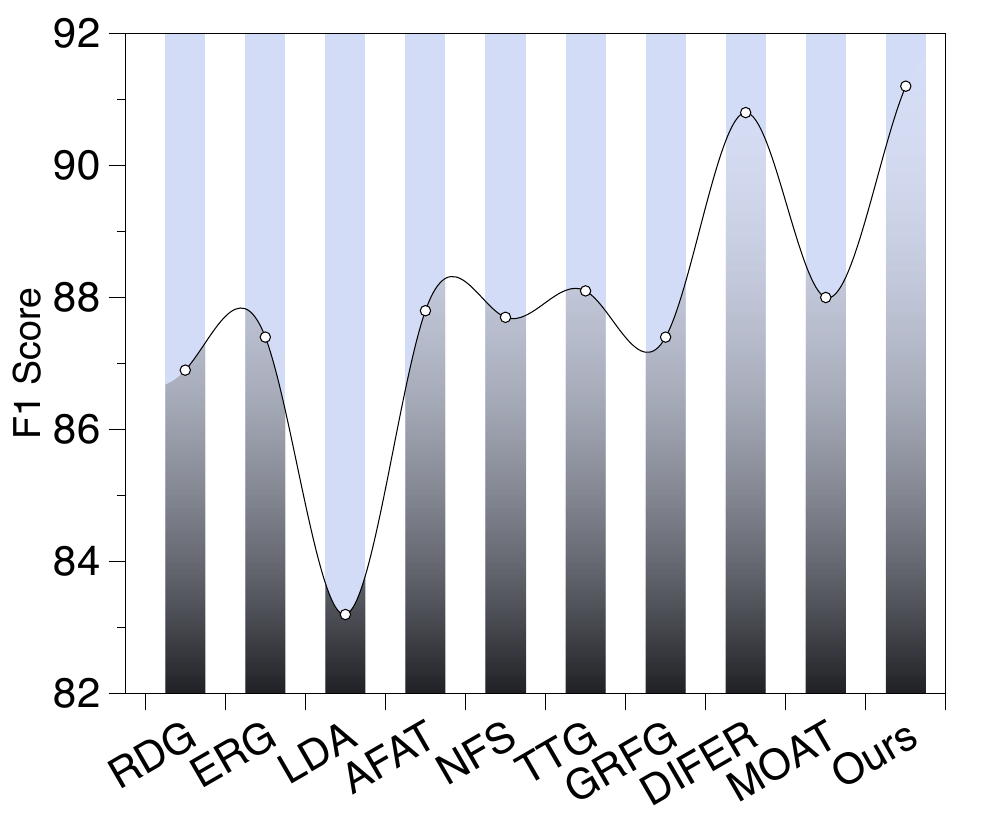}
    }
    \vspace{-0.4cm}
    \caption{Robustness check of \model\ with distinct ML models on SpamBase in terms of F1-score.}
    \label{robustness_check_fig}
    % \vspace{-0.65cm}
\end{figure*}

\begin{figure}
    \vspace{-0.6cm}
    \begin{center}
        \includegraphics[width=0.5\textwidth, trim={0.8cm 0 0 0}]{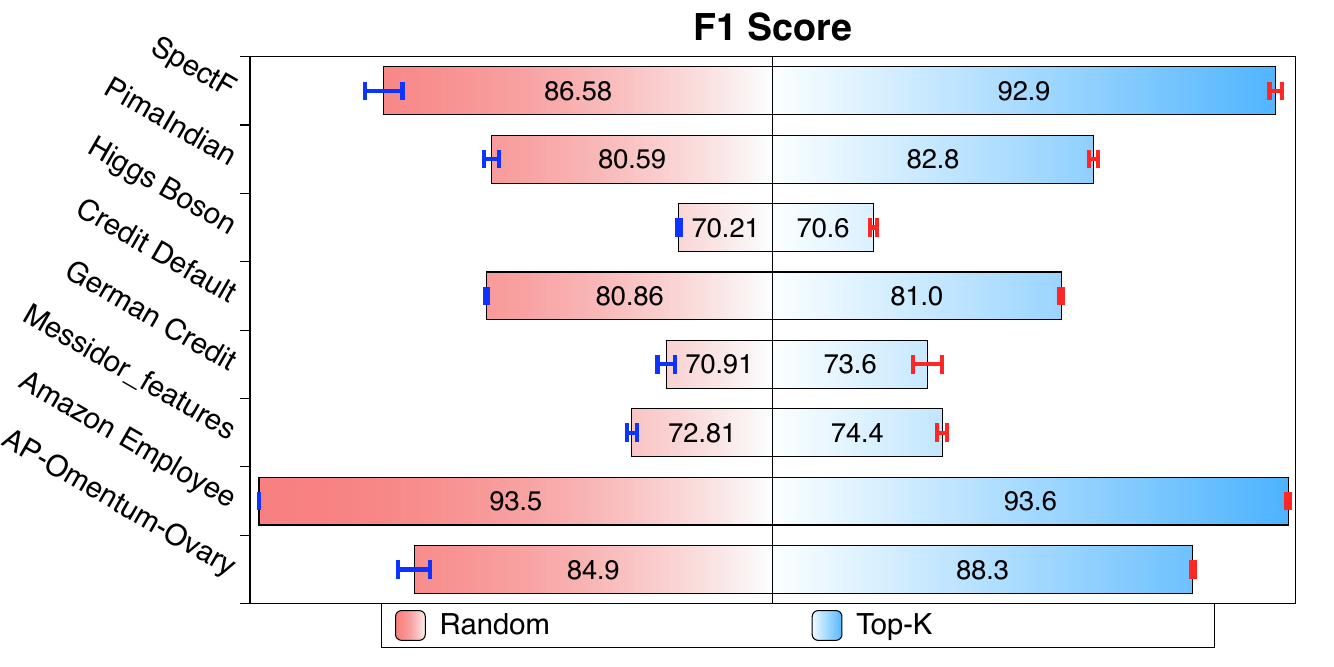}
    \end{center}
    % \vspace{-0.5cm}
    \caption{The influence of search seeds.}
    \vspace{0.4cm}
    \label{search_seed}
\end{figure}

\subsection{Performance Evaluation}
\subsubsection{Overall Performance.}
Table~\ref{main_table} reports the performance comparison between {\model} and baseline models across 19 datasets, using F1-score for classification tasks and 1-RAE for regression tasks.
All experimental results are obtained by independently running five times with different random seeds, showing that {\model} outperforms the other baseline models on all datasets.
There are three potential reasons for this observation:
1) The hierarchical modeling architecture captures both low-level and high-level relationships between features, encoding diverse feature transformation knowledge and building a more informative global embedding space;
2) The permutation-invariant concept encoder-decoder framework eliminates order sensitivity among concepts, constructing a global unbiased embedding space;
3) The policy-based RL agent effectively explores the global embedding space and identifies superior feature transformation embeddings, navigating the non-convex space more robustly than gradient-based methods and converging to higher-quality embedding regions.
Thus, this experiment demonstrates the effectiveness of our proposed framework in transforming feature spaces across various types of tasks.

% \vspace{-0.25cm}
% \vspace{0.25cm}
\subsubsection{Ablation Study.}
We design three model variants of {\model} to study the contribution of each component:
1) ${\model}^{-c}$ removes the concept encoder-decoder model;
2) ${\model}^{-p}$ replaces the concept encoder and decoder with permutation-sensitive self-attention layers;
3) ${\model}^{-g}$ replaces the policy-guided search with Genetic Algorithm (GA).
We select two classification tasks (SpectF, PimaIndian) and two regression tasks (Openml\_589, Openml\_620), evaluating each from four perspectives (F1-score, Precision, Recall, ROC/AUC for classification; 1-RAE, 1-MAE, 1-MSE, 1-RMSE for regression).
As shown in Figure~\ref{ablation}, {\model} consistently outperforms all variants, confirming that:
1) the hierarchical modeling module captures both low-level feature relationships and high-level concepts, preserving more informative feature transformation knowledge within the global embedding space;
2) the permutation-invariant mechanism removes permutation noise among generated concepts in the embedding space, facilitating more effective exploration by the RL agent.
3) the RL agent conducts effective exploration in the embedding space, eliminating the reliance on convexity assumptions and reducing the likelihood of convergence to local optima.
Thus, this experiment shows the significance of each component in {\model}.

\begin{figure}[!t]
    \vspace{-0.4cm}
    \begin{center}
        \subfigure[Sequence Length]{
            \includegraphics[width=0.235\textwidth]{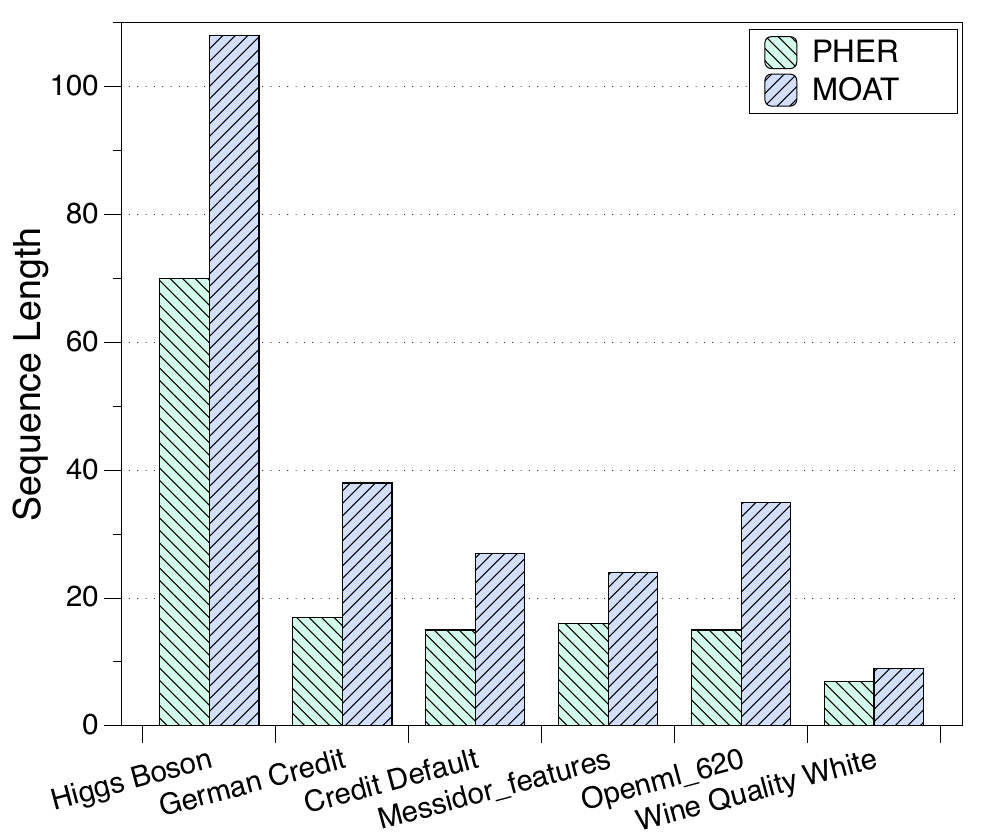}
        }
        \hspace{-4mm}
        \subfigure[Performance]{
            \includegraphics[width=0.235\textwidth]{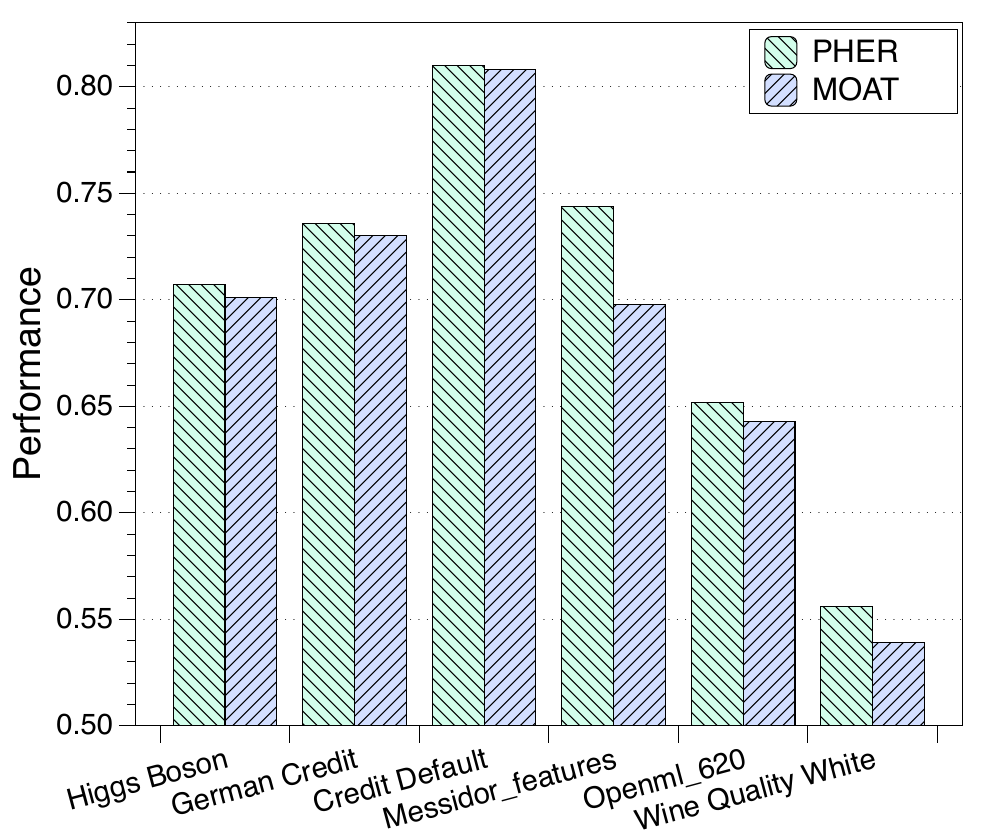}
        }
    \end{center}
    % \vspace{-0.5cm}
    \caption{Model scalability comparison between our method and the SOTA model MOAT.}
    \vspace{0.4cm}
    \label{model_scalability_fig}
\end{figure}
% \vspace{0.55cm}

\subsubsection{Robustness Check.}
\label{robustness_check}
We evaluate {\model} using various downstream ML models, including Random Forest (RF), XGBoost (XGB),  K-Nearest Neighborhood (KNN), and Decision Tree (DT), to assess the robustness of {\model}.
Figure~\ref{robustness_check_fig}  shows the comparison results on SpamBase in terms of the F1-score.
We observed that {\model} consistently outperforms all baseline methods across different downstream ML models.
A potential reason for this observation is that the RL-based data collector explores diverse feature transformation records guided by validation performance rather than any specific classifier.
These records further enable the hierarchical model to encode task-relevant knowledge and characteristics into the global embedding space.
Finally, the policy-guided agent leverages these records to effectively search the learned global embedding space, enabling the identification of high-quality feature transformation sequences that generalize well across different downstream classifiers.
Thus, this experiment evaluates the robustness of {\model}.
% The results on Support Vector Machine (SVM), Ridge, and LASSO are reported in Appendix~\ref{robustness_check_1}, Figure~\ref{robustness_check_1_fig}.

\begin{figure*}[!h]
    \centering
    \subfigure[SpectF]{
        \includegraphics[width=0.235\linewidth, trim = {0 0 1.5cm 0}]{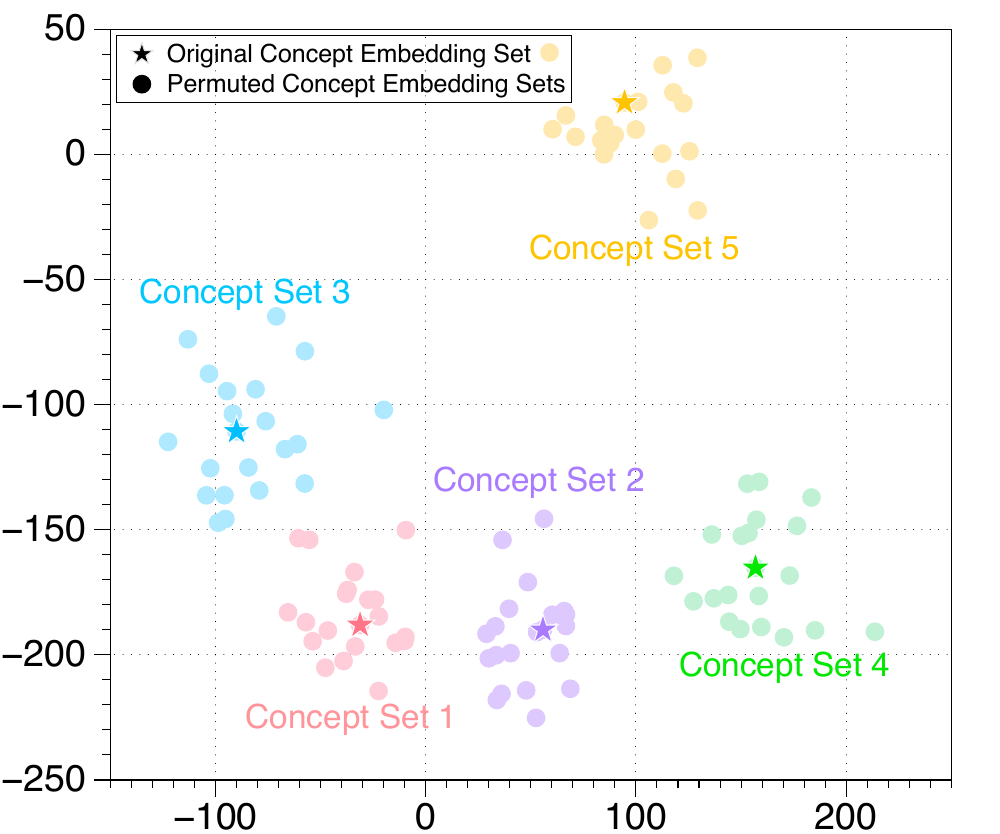}
    }
    \hspace{-1mm}
    \subfigure[German Credit]{
        \includegraphics[width=0.235\linewidth, trim = {0 0 1.5cm 0}]{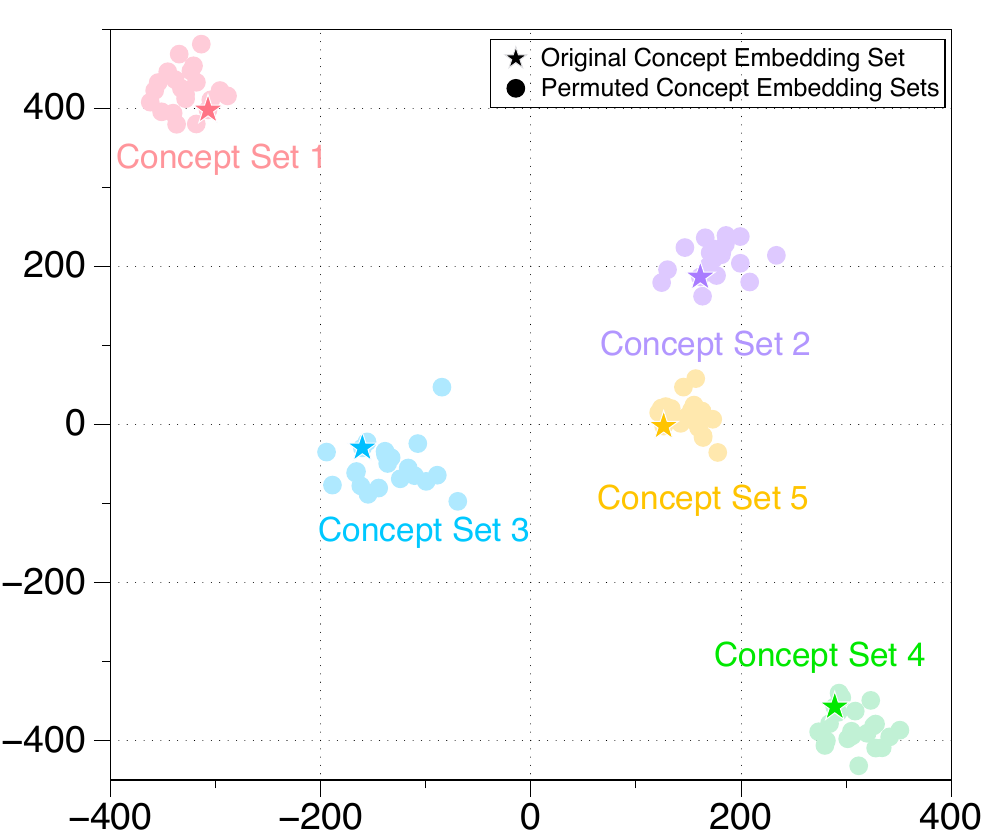}
    }
    \hspace{-1mm}
    \subfigure[SVMGuide3]{
        \includegraphics[width=0.235\linewidth, trim = {0 0 1.5cm 0}]{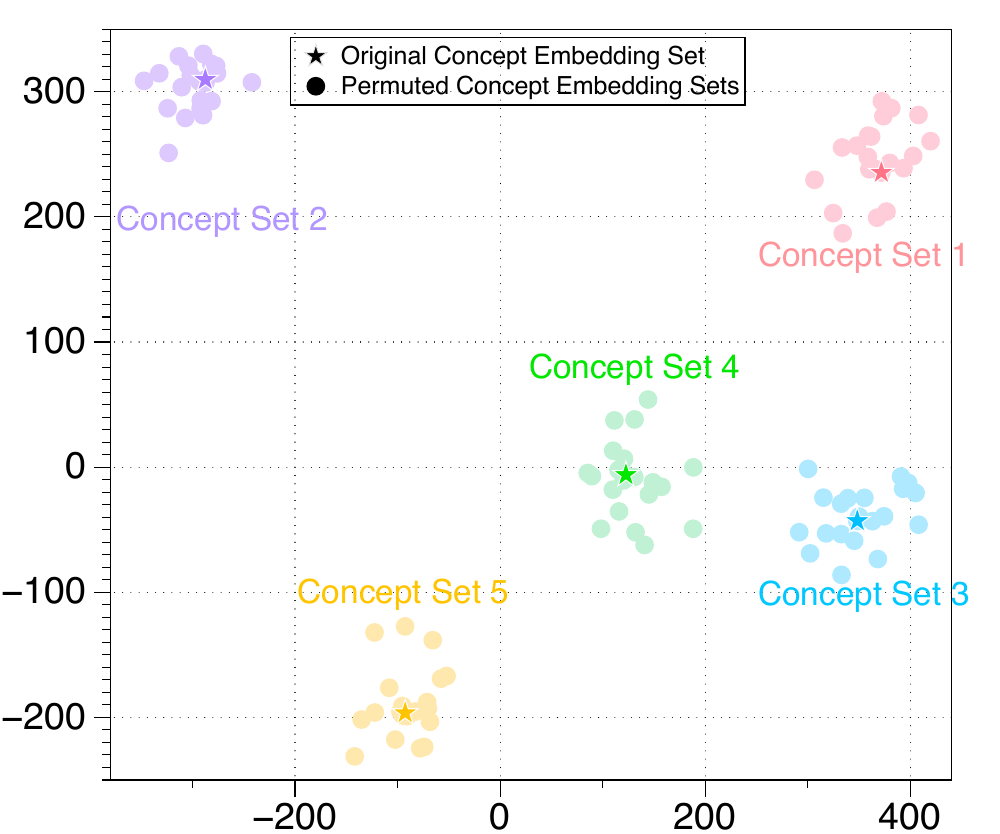}
    }
    \hspace{-1mm}
    \subfigure[Openml\_618]{
        \includegraphics[width=0.235\linewidth, trim = {0 0 1.5cm 0}]{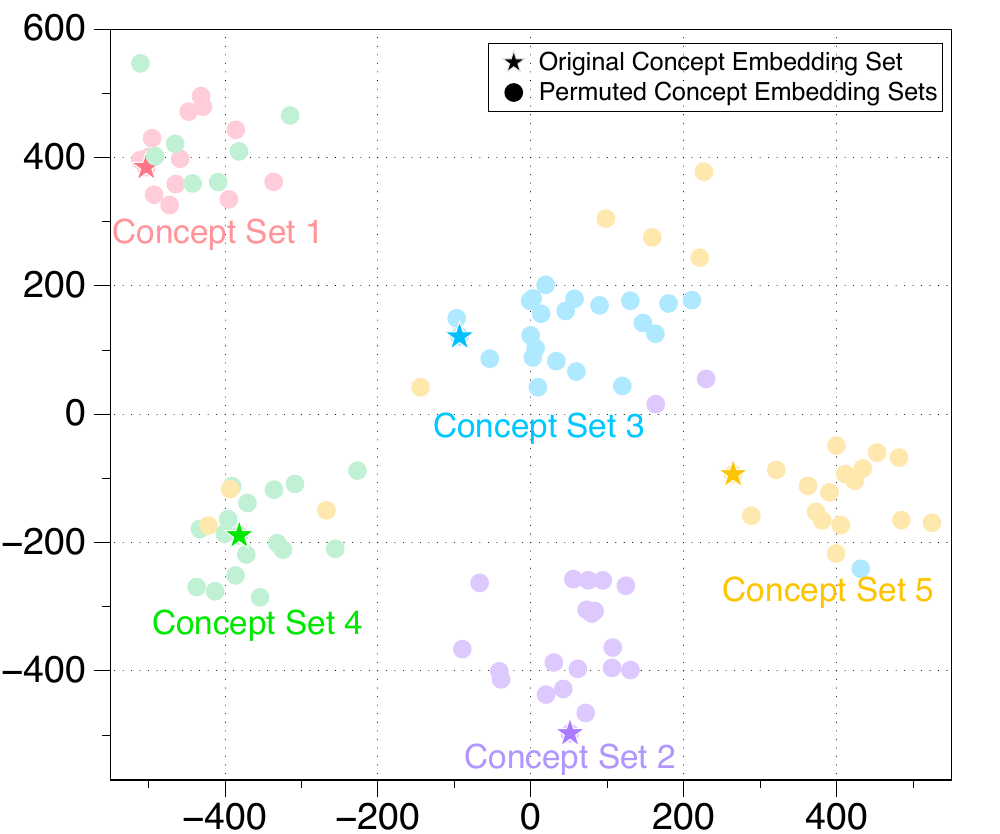}
    }
    \vspace{-0.2cm}
    \caption{Visualization of global embeddings obtained from original and permuted concept embeddings.}
    % \vspace{-0.3cm}
    \label{fig:permutation}
\end{figure*}

\begin{figure}
    % \vspace{-0.6cm}
    \begin{center}
        \subfigure[Clipping Trade-off]{
            \includegraphics[width=0.24\textwidth, trim=2cm 0.5cm 0cm 0.5cm, clip]{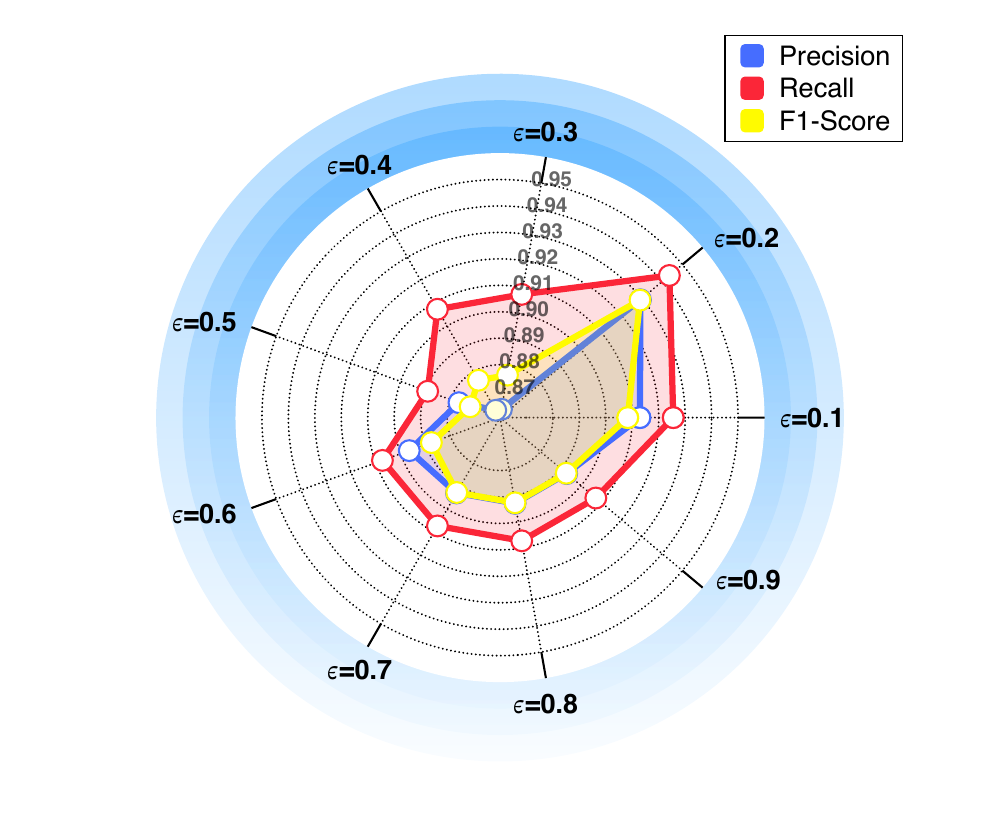}
        }
        \hspace{-6mm}
        \subfigure[Reward Trade-off]{
            \includegraphics[width=0.24\textwidth, trim=2cm 0.5cm 0cm 0.5cm, clip]{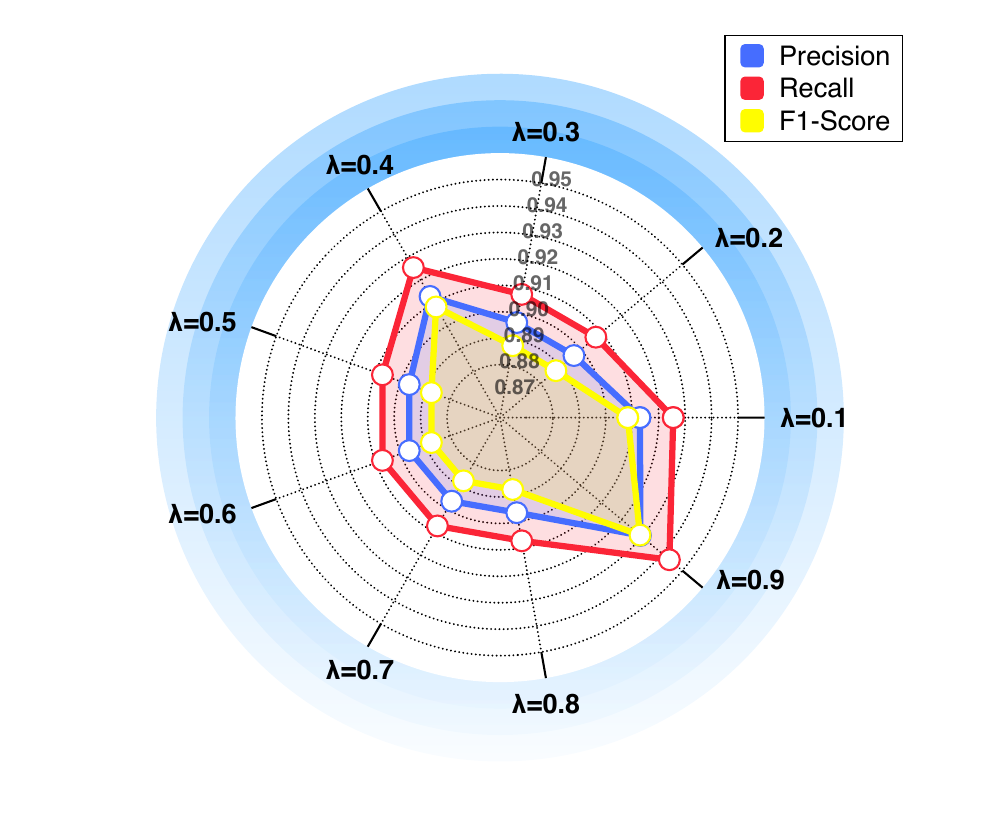}
        }
    \end{center}
    % \vspace{-0.3cm}
    \caption{Hyperparameter sensitivity.}
    \vspace{0.4cm}
    \label{hyper_fig}
\end{figure}

\subsubsection{Search Seeds.}
To observe the impacts of search seeds on the RL-based search process, we replace
the top K historical feature transformation records with K random records.
We conduct this experiment on eight randomly chosen datasets and report the average F1-score over five independent runs.
As shown in Figure~\ref{search_seed}, search initialized with high-quality seeds consistently outperforms search with random initialization.
A potential reason for this observation is that the RL agent can leverage informative seeds to anchor the search within promising regions of the embedding space, effectively narrowing the search scope and accelerating convergence.
In contrast, random starting points increase the risk of being trapped in local optima, resulting in an unstable search process and suboptimal feature transformation sequences.
Thus, this experiment demonstrates the significance of the search seeds in our framework.

\subsubsection{Model Scalability.}
\label{model_scalability}
To analyze model scalability, we compare the length of feature transformation sequences and downstream task performance of {\model} and the SOTA model MOAT.
From Figure~\ref{model_scalability_fig} (a), we observe that {\model} produces significantly shorter feature transformation sequences, showing its ability to discover compact and effective feature transformation sequences.
From Figure~\ref{model_scalability_fig} (b), we find that {\model} consistently outperforms MOAT across all datasets.
These observations suggest that the RL agent in {\model} effectively optimizes the learned embeddings by jointly maximizing task performance and minimizing transformation sequence length.
Thus, this experiment demonstrates that {\model} consistently achieves strong performance and compactness across diverse datasets, highlighting the scalability of {\model}.

\subsubsection{Permutation Sensitivity.}
\label{permutation_sensitivity}
We conduct this experiment to study the permutation sensitivity of learned global embeddings by visualizing embeddings derived from original and permuted concept sets on four datasets.
For each dataset, we randomly choose five samples as representatives and further permute the concept embedding of each sample 20 times.
Specifically, for each example, we first obtain its concept embedding set and create multiple permuted variants by randomly shuffling the order of the concept embeddings. Both the original and permuted concept embedding sets are then passed through the trained concept encoder to obtain their global embeddings. Thereafter, we visualize these embeddings using t-SNE.
Figure~\ref{fig:permutation} shows the visualization results, where each color represents the global embedding of one concept embedding set together with the embeddings derived from its permuted variants.
We find that the global embeddings derived from permuted concept orders consistently cluster around the original embedding.
This is because the self-attention pooling in $\phi_{con}$ computes a weighted aggregation over all concept embeddings, where the attention weights depend solely on the content similarity between fixed learnable seed queries and concept embeddings rather than their positions. Since weighted summation is a commutative operation, reordering the input concepts yields an identical global embedding, thereby eliminating permutation noise.
Therefore, this experiment confirms that our proposed hierarchical encoder-decoder effectively eliminates permutation bias, which is consistent with the theoretical property of the self-attention pooling mechanism.

\begin{figure}[!t]
    \centering
    % \vspace{-0.2cm}
    \subfigure[Original Feature Space]{
        \includegraphics[width=0.22\textwidth]{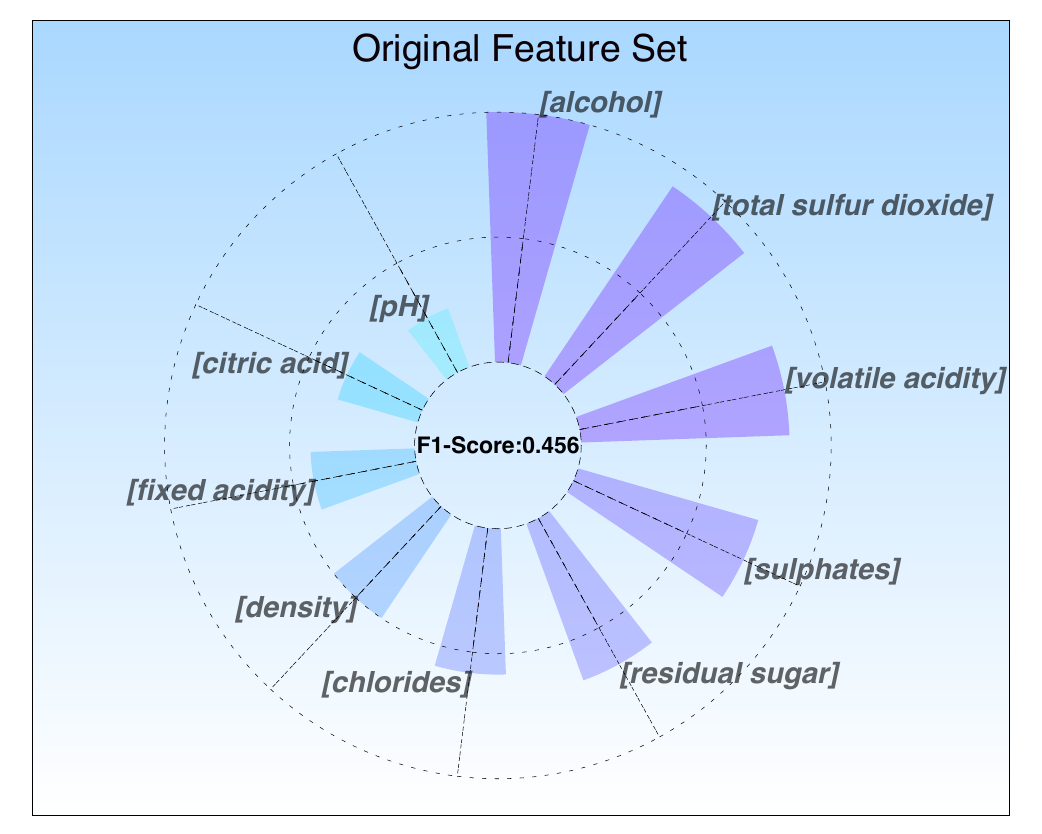}
    }
    % \hspace{-3mm}
    \subfigure[{\model} Generated Feature Space]{
        \includegraphics[width=0.22\textwidth]{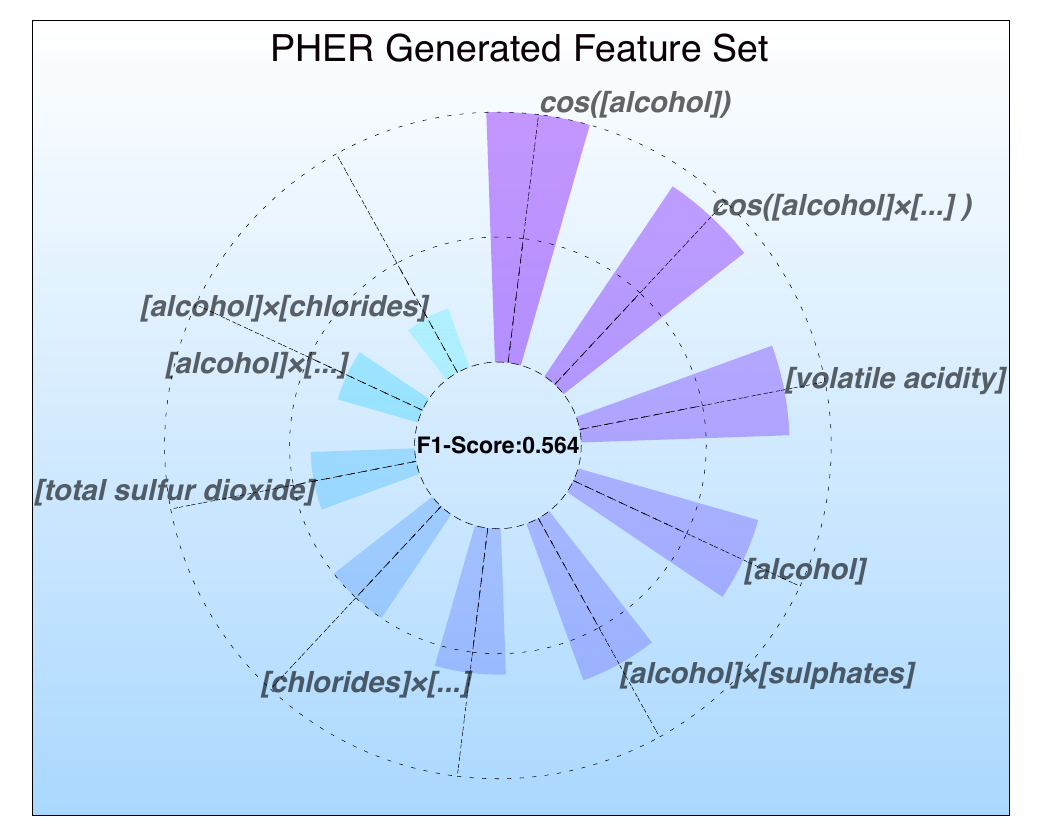}
    }
    % \hspace{-4mm}
    \vspace{0.3cm}
    \caption{Comparison of traceability on the original feature set and selected feature subset.}
    \vspace{0.4cm}
    \label{case_study_fig}
\end{figure}

\subsubsection{Hyperparameter Sensitivity Analysis.}
We assess the hyperparameter sensitivity of {\model} on the SpectF dataset by varying the clipping ratio $\epsilon$ and the reward trade-off $\lambda$ from 0.1 to 0.9, where $\epsilon$ stabilizes training by controlling the magnitude of policy updates and $\lambda$ balances task performance against transformation sequence length.
Figure~\ref{hyper_fig} demonstrates the overall experimental results in terms of Precision, Recall and F1-Score.
We found that {\model} achieves optimal performance
when $\epsilon=0.2$ and $\lambda=0.9$.
This aligns with prior PPO studies, where $\epsilon=0.2$ is a setting widely adopted in standard reinforcement learning benchmarks~\cite{ppo}.
Furthermore, we found that the model performance remains relatively stable across a broad range of $\lambda$, with slight improvements as $\lambda$ approaches 0.9.
Therefore, this experiment demonstrates the sensitivity of our framework to hyperparameter settings and provides practical guidance on hyperparameter configuration.

\subsubsection{Traceability Case Study}
\label{case_study}
We conduct this experiment to evaluate the traceability of {\model}.
We rank the top 10 most significant features for prediction in both the original feature set and {\model} generated feature set of the Wine Quality Red dataset.
Figure~\ref{case_study_fig} visualizes the experiment results, where a larger bar indicates higher feature importance.
We observed that approximately 70\% of the crucial features in the new feature set are generated by {\model}.
The newly generated feature space enhances the downstream ML performance by 23.7\%.
There are two potential reasons for this observation:
1) the hierarchical modeling module in {\model} effectively captures the inherent hierarchical relationships from low-level features and operations to high-level concepts.
2) the policy-guided multi-objective search strategy effectively explores the learned global embedding space and identifies the superior feature transformation sequence, overcoming the non-convex challenge and converging to the global optimal embedding point.
Furthermore, we found that `[alcohol]’ is the most important feature in the original set.
This aligns with domain knowledge, as alcohol is known to be one of the most influential factors in determining red wine quality.
    {\model} not only identifies this essential feature but also generates a variety of composited features based on `[alcohol]’, which further enhance the predictive performance.
This observation demonstrates that {\model} is capable of identifying the importance of individual features while also deriving new informative representations that align with domain semantics and enhance downstream performance.
Such new informative features empower domain experts to trace the origins of transformed features and derive novel analytical rules for assessing red wine quality.
Thus, this case study shows the traceability and interpretability of {\model}.

\section{Related Works}
\textbf{Automated Feature Transformation (AFT)} aims to enhance the tabular feature space by systematically applying mathematical operations to original features, thereby improving feature expressiveness and downstream predictive performance~\cite{chen2021techniques,kusiak2001feature}. 
Existing methods can be broadly categorized into three main classes.
1) expansion-reduction based approaches~\cite{kanter2015deep,khurana2016cognito, lam2017one,afat,khurana2016automating} first expand the feature space by enumerating candidate features through predefined mathematical transformations, and subsequently reduce dimensionality via feature selection or pruning strategies.
While effective for capturing simple feature interactions, these methods rely on shallow compositions and heuristic selection criteria, making it difficult to represent complex, multi-step feature transformations, which often leads to suboptimal performance in practice.
2) evolution-evaluation approaches~\cite{grfg, ttg,tran2016genetic,zhu2022evolutionary, xiao2022traceable,xiao2023traceable} 
integrate feature generation and selection into a closed-loop optimization framework.
By leveraging evolutionary algorithms or reinforcement learning (RL), these methods iteratively explore transformation operators and retain features that improve validation performance.
Despite their flexibility, such approaches typically suffer from high computational cost and unstable optimization behavior, largely due to discrete decision-making and the combinatorial nature of the transformation space.
3) Auto ML-based approaches~\cite{elsken2019neural,li2021automl, he2021automl, karmaker2021automl,zhang2021automated,wever2021automl,bahri2022automl,wang2021autods,xiao2023discrete,ying2023self,ren2023mafsids} formulate AFT as part of a broader AutoML pipeline, jointly searching for feature transformation strategies and model configurations.
A representative example is~\cite{moat}, which embeds RL-collected feature transformation sequences into a continuous representation using postfix expressions and applies gradient-ascent beam search to identify informative transformation embeddings.
However, such methods are limited by:
1) overlooking intricate hierarchical relationships inherent in feature transformation knowledge;
2) encoding feature transformation sequences as order-sensitive; 
3) relying on the convexity assumption of the embedding space.
To address these drawbacks, {\model} combines permutation-invariant hierarchical modeling and multi-objective policy-guided search.
The permutation-invariant hierarchical modeling module captures both token-level and concept-level feature transformation knowledge and mitigates order sensitivity among concepts, creating an unbiased global embedding space.
The multi-objective policy-guided search effectively explores the learned embedding space, identifying better feature transformation sequences without relying on any convexity assumptions.

% \noindent\textbf{Comparison with Prior Literature.} 
% While existing approaches have made substantial progress, current formulations generally operate at the feature level without capturing the inherent hierarchical structure in transformation processes. 
% Moreover, the resulting embedding space is typically order-sensitive and optimized through gradient-based search method.
% In contrast, {\model} incorporates hierarchical modeling to jointly preserve token-level interactions and concept-level abstractions and further employs a permutation-invariant mechanism to remove order-induced biases. 
% Thereafter, we adopt a  multi-objective policy-guided search strategy to explore the non-convex embedding space.

\vspace{-0.2cm}
\section{Conclusion Remarks}
\label{conclusion}
In this paper, we propose a hierarchical feature transformation framework {\model} that integrates permutation-invariant hierarchical modeling and multi-objective policy-guided search.
In detail, we first develop a permutation-invariant hierarchical modeling module, including token-level and concept-level encoder-decoder models, to preserve feature transformation knowledge at both the feature--operation token level and the generated concept level into a global embedding space.
Within this module, we develop a self-attention pooling mechanism that symmetrically computes attention scores across all generated concepts to ensure permutation invariance.
Then, we employ a multi-objective search strategy to explore the learned embedding space, overcoming the reliance on convexity assumptions and mitigating the risk of being trapped in local optima.
Finally, extensive experiments demonstrate several key insights: 
1) the hierarchical modeling structure significantly captures meaningful hierarchical interactions, enhancing the expressivity of the learned embedding space.
2) the permutation-invariant module effectively mitigates order sensitivity, stabilizing the embedding space learning and search processes. 
3) the policy-guided RL search enables effective exploration, mitigating the risk of convergence to local optima and improving search robustness.
These findings highlight the importance of permutation-invariant hierarchical modeling and robust exploration strategies for advancing automated feature transformation. 
For future research, a promising direction is to improve the computational efficiency and scalability of {\model}, possibly by integrating a lightweight concept modeling framework or refining the RL-based search strategy in the embedding space.

\section*{GenAI Usage Disclosure}
Generative AI tools were used only for language polishing, grammar refinement, and auxiliary code assistance, including code editing and debugging suggestions.
All research ideas, experimental designs, code, data processing, analyses, and final manuscript content were created, verified, and approved by the authors.

% \input{acknowledgment}


% \newpage
\bibliographystyle{ACM-Reference-Format}
\balance
\bibliography{sample-base}

\clearpage

\end{document}